%% file: arxiv.tex
\documentclass[10pt,twocolumn,letterpaper]{article}

\usepackage{3dv}
\usepackage{times}
\usepackage{epsfig}
\usepackage{graphicx}
\usepackage{amsmath}
\usepackage{amssymb}

\usepackage{afterpage}
\usepackage{amsmath}
\usepackage{amssymb}
\usepackage{array}
\usepackage{bm}
\usepackage{booktabs}
\usepackage{enumitem}
\usepackage{fancyhdr}
\usepackage{float}
\usepackage{graphicx}
\usepackage{listings}
\usepackage{multirow}
\usepackage{subcaption}

\usepackage{pifont}

\newcommand{\stufig}[5]                                    
{
	\begin{figure}[#5]
		\begin{center}
			\includegraphics[#1]{#2}
			\caption{#3}
			\label{#4}
		\end{center}
		\vspace{-.7cm}
	\end{figure}
}

\newcommand{\stufigstar}[5]                                
{
	\begin{figure*}[#5]
		\begin{center}
			\includegraphics[#1]{#2}
			\caption{#3}
			\label{#4}
		\end{center}
		\vspace{-.7cm}
	\end{figure*}
}

\newenvironment{stusubfig}[1]
{
	\begin{figure}[#1]
		\begin{center}
		}
		{
		\end{center}
	\end{figure}
}

\newenvironment{stusubfig*}[1]
{
	\begin{figure*}[#1]
		\begin{center}
		}
		{
		\end{center}
	\end{figure*}
}

\newcolumntype{H}{>{\setbox0=\hbox\bgroup}c<{\egroup}@{}}

\setlist{noitemsep,topsep=0pt,parsep=0pt,partopsep=0pt}

\usepackage[pagebackref=false,breaklinks=true,letterpaper=true,colorlinks,bookmarks=false]{hyperref}

\threedvfinalcopy 

\def\threedvPaperID{***} 

\ifthreedvfinal\fi

\begin{document}

\title{Let's Take This Online: Adapting Scene Coordinate Regression \\ Network Predictions for Online RGB-D Camera Relocalisation}

\author{
\begin{tabular}{c}
\begin{tabular}{c@{\hskip 0.85cm}c@{\hskip 0.85cm}c}
Tommaso Cavallari$^{1,}$\thanks{Authors contributed equally.} & Luca Bertinetto$^1$ & Jishnu Mukhoti$^1$
\end{tabular}
\\
\begin{tabular}{c@{\hskip 0.85cm}c}
Philip Torr$^{1,2}$ & Stuart Golodetz$^{1,*}$
\end{tabular}
\\\\
\begin{tabular}{c}
$^1$ Oxford Research Group, FiveAI Ltd. \\
$^2$ Department of Engineering Science, University of Oxford
\end{tabular}
\\
\begin{tabular}{c}
\normalsize\texttt{\{tommaso.cavallari,luca.bertinetto,jishnu.mukhoti,stuart\}@five.ai} \\
\normalsize\texttt{philip.torr@eng.ox.ac.uk}
\vspace{-.8\baselineskip}
\end{tabular}
\end{tabular}
}

\maketitle

\begin{abstract}
\noindent Many applications require a camera to be relocalised online, without expensive offline training on the target scene. Whilst both keyframe and sparse keypoint matching methods can be used online, the former often fail away from the training trajectory, and the latter can struggle in textureless regions. By contrast, scene coordinate regression (SCoRe) methods generalise to novel poses and can leverage dense correspondences to improve robustness, and recent work has shown how to adapt SCoRe forests between scenes, allowing their state-of-the-art performance to be leveraged online. However, because they use features hand-crafted for indoor use, they do not generalise well to harder outdoor scenes. Whilst replacing the forest with a neural network and learning suitable features for outdoor use is possible, the techniques used to adapt forests between scenes are unfortunately harder to transfer to a network context. In this paper, we address this by proposing a novel way of leveraging a network trained on one scene to predict points in another scene. Our approach replaces the appearance clustering performed by the branching structure of a regression forest with a two-step process that first uses the network to predict points in the original scene, and then uses these predicted points to look up clusters of points from the new scene. We show experimentally that our online approach achieves state-of-the-art performance on both the 7-Scenes and Cambridge Landmarks datasets, whilst running in under 300ms, making it highly effective in live scenarios.
\end{abstract}

\input{text/introduction}
\input{text/method}
\input{text/experiments}
\input{text/conclusion}

\part*{Supplementary Material}

\appendix

\input{text/supp-arxiv}

\stufig{width=\linewidth}{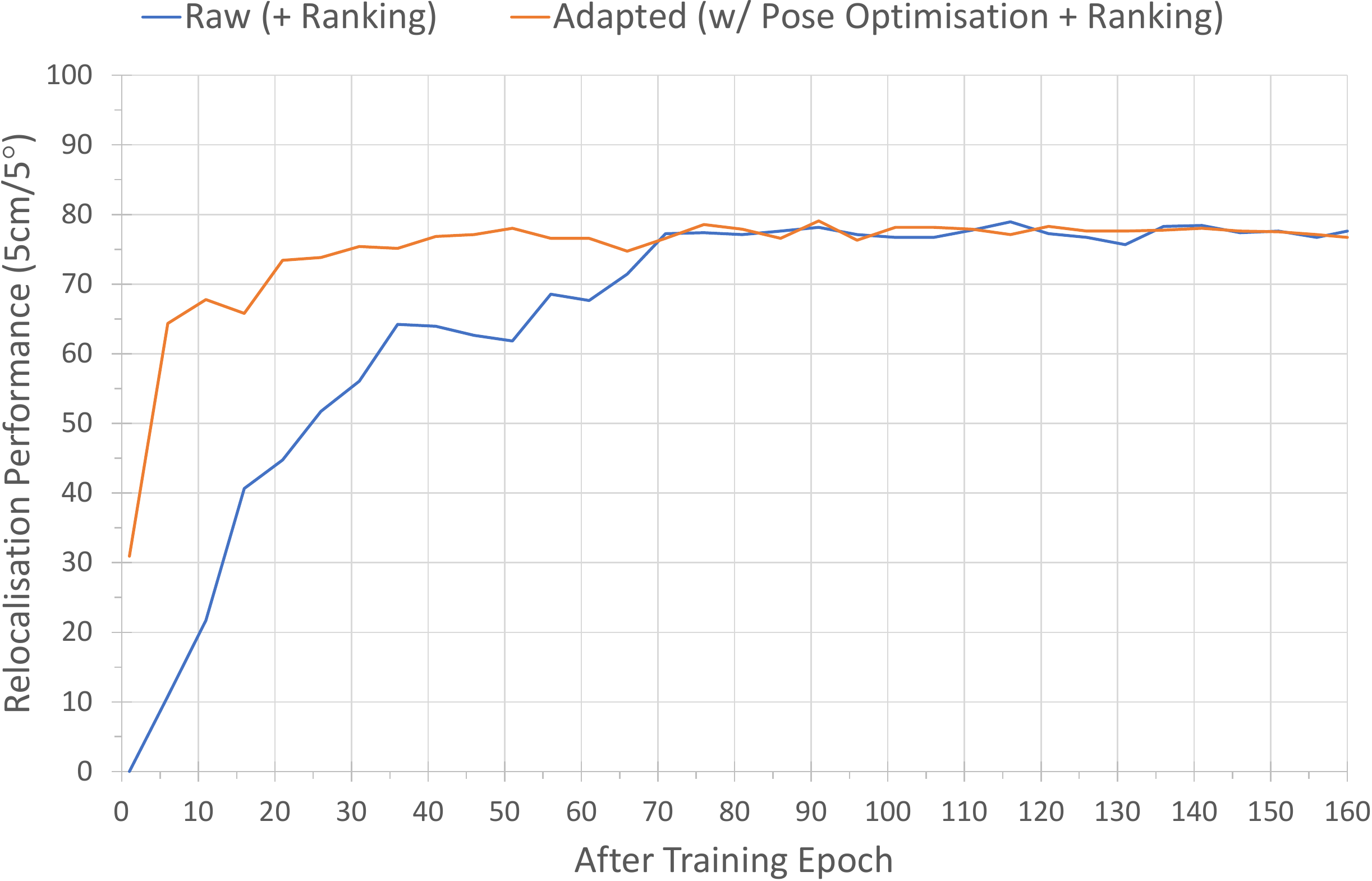}{Evaluating how the relocalisation performance (after hypothesis ranking, both with/without grid-based adaptation with $\ell = 4$m) of a ScoreNet trained on \emph{Great Court} \cite{Kendall2015,Kendall2016,Kendall2017}, and tested on \emph{Great Court} after every $5$ training epochs, varies during the training process. In both cases, we are able to start relocalising relatively well after less than $40$ epochs, and there is little additional gain in performance after around $70$ epochs.}{fig:training-efficiency}{!t}

\section*{Acknowledgements}

\noindent This work was supported by FiveAI Ltd.\ and Innovate UK/CCAV project 103700 (StreetWise). We are grateful to the authors of \cite{Brachmann2018CVPR} for sharing their depth images for the Cambridge Landmarks dataset with us.

{
\small
\bibliographystyle{ieee}
\bibliography{arxiv}
}

\newpage

\input{text/supptables}

\end{document}

%% file: text/introduction.tex
\vspace{-\baselineskip}

\section{Introduction}

\noindent Visual-only camera relocalisation has a wide variety of applications across computer vision and robotics, including augmented reality \cite{Castle2008,Paucher2010,Golodetz2015SPTR,Rodas2015,Bae2016}, tracking recovery and loop closure during SLAM \cite{Williams2011,MurArtal2014,Cavallari2019}, and map merging \cite{Kaehler2016,Golodetz2018}. For many applications, an ability to relocalise a camera \emph{online}, i.e.\ without expensive prior training on the scene of interest, is critical. For example, in an interactive SLAM context, it is typical to initialise the pose of the camera at the start of reconstruction and then track it from one frame to the next,
but when that tracking inevitably fails at some point, it is important to be able to relocalise the camera as soon as possible so that reconstruction can continue without unnecessary delay. However, despite the significant research attention that has been devoted to camera relocalisation in recent years, many state-of-the-art methods (especially those based on regression) remain wedded to an offline setting, making them difficult to deploy for live use.
Existing methods can be broadly divided into five types:

(i) \emph{Global matching} methods match one or more frames (or a descriptor, point cloud or 3D model derived from them) against either the contents of a database (e.g.\ one containing a map from keyframes to known poses), or a model of the scene, to look up a suitable pose. Methods like \cite{Gee2012} match the image itself against synthetic views of the scene, whereas \cite{GalvezLopez2011} and \cite{Glocker2015} match image descriptors against a database. Other methods \cite{Laskar2017,Balntas2018} find the nearest neighbours to a query image in the database, and use their poses to determine the query pose. Such image retrieval methods can struggle to generalise to novel poses. Geometry-based matching methods avoid this, but often require more than a single frame from which to relocalise. For example, \cite{Deng2016} matches a point cloud constructed from a set of query images to a point cloud of the scene, whilst \cite{Lu2016} reconstructs a 3D model from a short video sequence and matches that against the scene. One exception is \cite{Schoenberger2018},
which matches hallucinated subvolumes against a database using a variational encoder-decoder network.
This is single-frame, but quite slow, taking around a second per frame to relocalise.

\stufigstar{width=.95\linewidth}{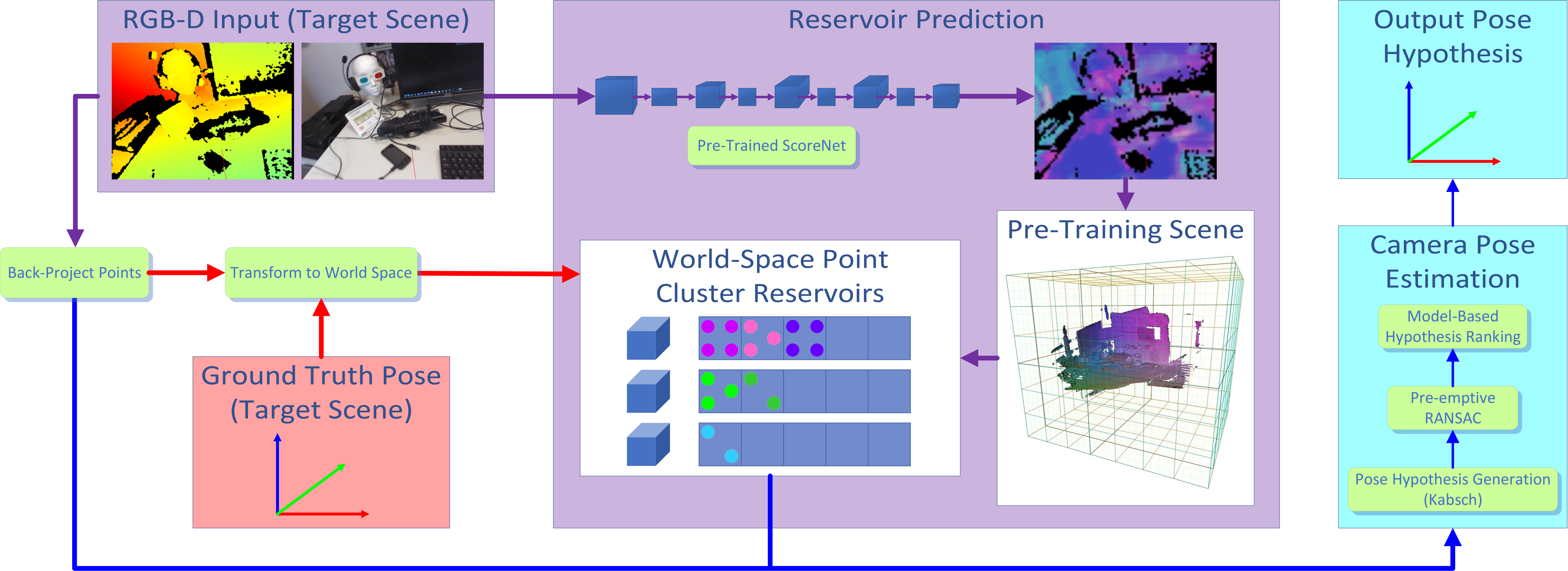}{\textbf{An overview of our approach.} Ahead of time, we train a scene coordinate regression network \emph{offline} to predict correspondences between pixels in an input image and 3D points in an arbitrary pre-training scene (here, \emph{Chess} \cite{Shotton2013}): see \S\ref{subsec:method-offlinetraining}. To use this network to predict points in a different target scene (here, \emph{Heads} \cite{Shotton2013}) \emph{online}, we use the points the network predicts to index into an array of reservoirs, implicitly clustering the pixels using a grid-based approach that uses predicted pre-training scene location as a proxy for appearance: see \S\ref{subsec:method-onlineadaptation}. At online training time (purple and red boxes), we fill the reservoirs with points from the target scene, which we cluster using Really Quick Shift \cite{Chum2003}. At test time (purple and blue boxes), we predict a reservoir for each pixel, and use the point clusters the reservoirs contain to generate correspondences that can be passed to a Kabsch-RANSAC camera pose estimation backend \cite{Cavallari2019} to relocalise the camera: see \S\ref{subsec:method-cameraposeestimation}.}{fig:pipeline}{!t}

(ii) \emph{Global regression} methods directly regress an image's pose, using e.g.\ decision forests \cite{Kacete2017}, pose regression networks \cite{Kendall2015,Kendall2016,Kendall2017,Melekhov2017,Wu2017,Acharya2019}, GANs \cite{Bui2019} or LSTMs \cite{Clark2017,Walch2017}. Various recent approaches \cite{Brahmbhatt2018,Radwan2018,Valada2018,Li2019} have made use of the relative poses between images to improve performance. Global regression methods have proved popular, but have typically struggled to achieve the accuracy of local methods (see below). Those methods that do achieve better performance \cite{Radwan2018,Valada2018} currently do so by relying on an estimated pose from the previous frame, thereby essentially performing camera tracking rather than single-image relocalisation. Indeed, recent work by Sattler et al.\ \cite{Sattler2019} has shown that global regression is in many ways conceptually similar to image retrieval, and that current such approaches do not consistently outperform an image retrieval baseline or generalise well to novel poses. (This link with image retrieval was also noted earlier by Wu et al.\ \cite{Wu2017}.)

(iii) \emph{Local matching} methods match points in camera space with known points in world space, pass the correspondences to the Perspective-n-Point (PnP) algorithm \cite{Hartley2004} or the Kabsch algorithm \cite{Kabsch1976} to generate a number of initial camera pose hypotheses, and then refine these down to a final pose using some variant of RANSAC \cite{Fischler1981}. Many approaches match the descriptors of sparse keypoints to perform this matching \cite{Williams2011,Li2015,Feng2017,Sattler2017}; some approaches that perform dense matching also exist \cite{Schmidt2017}. Local matching methods tend to generalise better to novel poses than image retrieval methods, since individual points are often easier to match from novel angles than are whole images.

(iv) \emph{Local regression} methods generally use regression forests \cite{Shotton2013,GuzmanRivera2014,Valentin2015RF,Brachmann2016,Meng2016,Cavallari2017,Meng2017IROS,Meng2018IROS,Cavallari2019}, neural networks \cite{Brachmann2017CVPR,Brachmann2018CVPR,Duong2018,Li2018RSS,Li2018ECCV,Brachmann2019}, or a mix of the two \cite{Massiceti2017} to predict the scene coordinates of pixels in the input image. They then pass these correspondences to PnP/Kabsch and RANSAC. Compared to local matching methods, local regression methods can avoid the need for explicit keypoint detection, which can be costly, and can make use of correspondences from the whole image during RANSAC, which can help with robustness \cite{Shotton2013}. Like local matching methods, they also generalise well from novel poses. However, whilst they tend to be more accurate than local matching methods at small/medium scale, they have not yet been shown to scale well to very large scenes \cite{Sattler2019}.

(v) \emph{Hybrid} methods use both the global and local paradigms, generally by first performing some kind of lookup/matching, and then refining the results using either RANSAC \cite{MurArtal2015,Taira2018} or continuous pose optimisation \cite{Valentin2016}.

Not all of these methods are designed for online, single-frame relocalisation. Image retrieval methods can normally be used online, but struggle to relocalise from novel poses. Global regression methods generally require significant offline training on the target scene; moreover, their comparatively poor accuracy makes them unattractive for applications like interactive dense SLAM \cite{Newcombe2011} that require precise poses (their main niche is large-scale, RGB-only relocalisation scenarios in which coarse poses are acceptable).
Local matching methods can generally be used online \cite{Williams2011,Feng2017}, but because most rely on detecting/matching sufficient sparse keypoints in the image, their robustness can suffer in textureless parts of the scene.
By contrast, local regression methods avoid the need to detect keypoints explicitly, making them appealing for robust relocalisation in small/medium-scale scenes. However, most such methods, like their global counterparts, require costly offline training.

One \emph{online} local regression approach is that of \cite{Cavallari2017,Cavallari2019}, which showed how to adapt the regression forests of \cite{Shotton2013} for online use in real time.
Their approach achieves state-of-the-art performance on the popular 7-Scenes \cite{Shotton2013} and Stanford 4 Scenes \cite{Valentin2016} indoor datasets, and also performs well on some of the easier outdoor scenes from Cambridge Landmarks \cite{Kendall2015,Kendall2016,Kendall2017}. However, because their forests use hand-crafted features that were designed for indoor use \cite{Shotton2013}, they struggle \cite{Cavallari2019} to work out-of-the-box on harder outdoor scenes.
Whilst it might in principle be possible to solve this problem by hand-crafting new features for outdoor use, doing so could be time-consuming and costly. Indeed, the broader trend in machine learning has been towards replacing models such as regression forests with neural networks that can learn suitable features, rather than trying to hand-craft them manually. However, replacing the forests used by \cite{Cavallari2017,Cavallari2019} with networks is not straightforward. To achieve online relocalisation, they rely on the way in which their forests predict leaves containing reservoirs of points to adapt forests between scenes, and it is tricky to see how this scheme can be easily transferred to work with local regression networks, which tend to directly predict individual points in the training scene.

\textbf{Contribution.} In this paper, we address this problem by proposing a novel method that allows the predictions of a network trained to regress 3D points in one scene to be leveraged to predict points in a new scene, thereby enabling the network to be used online. Our approach (see \S\ref{sec:method}) works by replacing the appearance clustering that was implicitly being performed by the branching structure of the forests in \cite{Cavallari2017,Cavallari2019} with a two-step process that first uses the network to predict points in the scene on which it was trained, and then uses these predicted points to look up reservoirs of points from the new scene. We show via experiments on 7-Scenes \cite{Shotton2013} and Cambridge Landmarks \cite{Kendall2015,Kendall2016,Kendall2017} that our approach achieves state-of-the-art performance in under $300$ms, whilst requiring no offline training on the test scene. We further show that the learnt features of our networks allow us to perform well even on harder outdoor scenes that
were causing methods such as that of \cite{Cavallari2017,Cavallari2019} to fail.

%% file: text/method.tex
\section{Method}
\label{sec:method}

\subsection{Overview}

\noindent Our pipeline is shown in Figure~\ref{fig:pipeline}. We start by training (\emph{offline}) a scene coordinate regression network (a `ScoreNet') to predict correspondences between pixels in an input image and 3D points in an arbitrary pre-training scene. The structure of the ScoreNets we use and how they are trained are described in \S\ref{subsec:method-offlinetraining}. To use a ScoreNet to relocalise in a scene other than the one on which it was trained, we need a way of adapting the predictions of the network \emph{online} so as to predict points in the new scene of interest. As described in more detail in \S\ref{subsec:method-onlineadaptation}, we do this by using the points predicted by the network to index into an array of reservoirs that can be refilled with points from the new scene at online training time, and then looked up again at test time to generate correspondences. This scheme draws inspiration from the adaptive regression forest approach of Cavallari et al.\ \cite{Cavallari2017,Cavallari2019}, but modifies it to work in a ScoreNet context. Having obtained the needed correspondences, we can then generate camera pose hypotheses using the Kabsch algorithm \cite{Kabsch1976} and refine them down to a single output pose using pre-emptive RANSAC, as was done in \cite{Cavallari2017}.
ICP \cite{Besl1992} against a 3D model of the scene can then be used to refine the initial pose produced by the relocaliser. Furthermore, the last few candidates considered by RANSAC can also be ranked using the model for an additional boost in performance, as described in \cite{Cavallari2019}. See \S\ref{subsec:method-cameraposeestimation} for more details.

\subsection{Offline ScoreNet Training}
\label{subsec:method-offlinetraining}

\stufig{width=.9\linewidth}{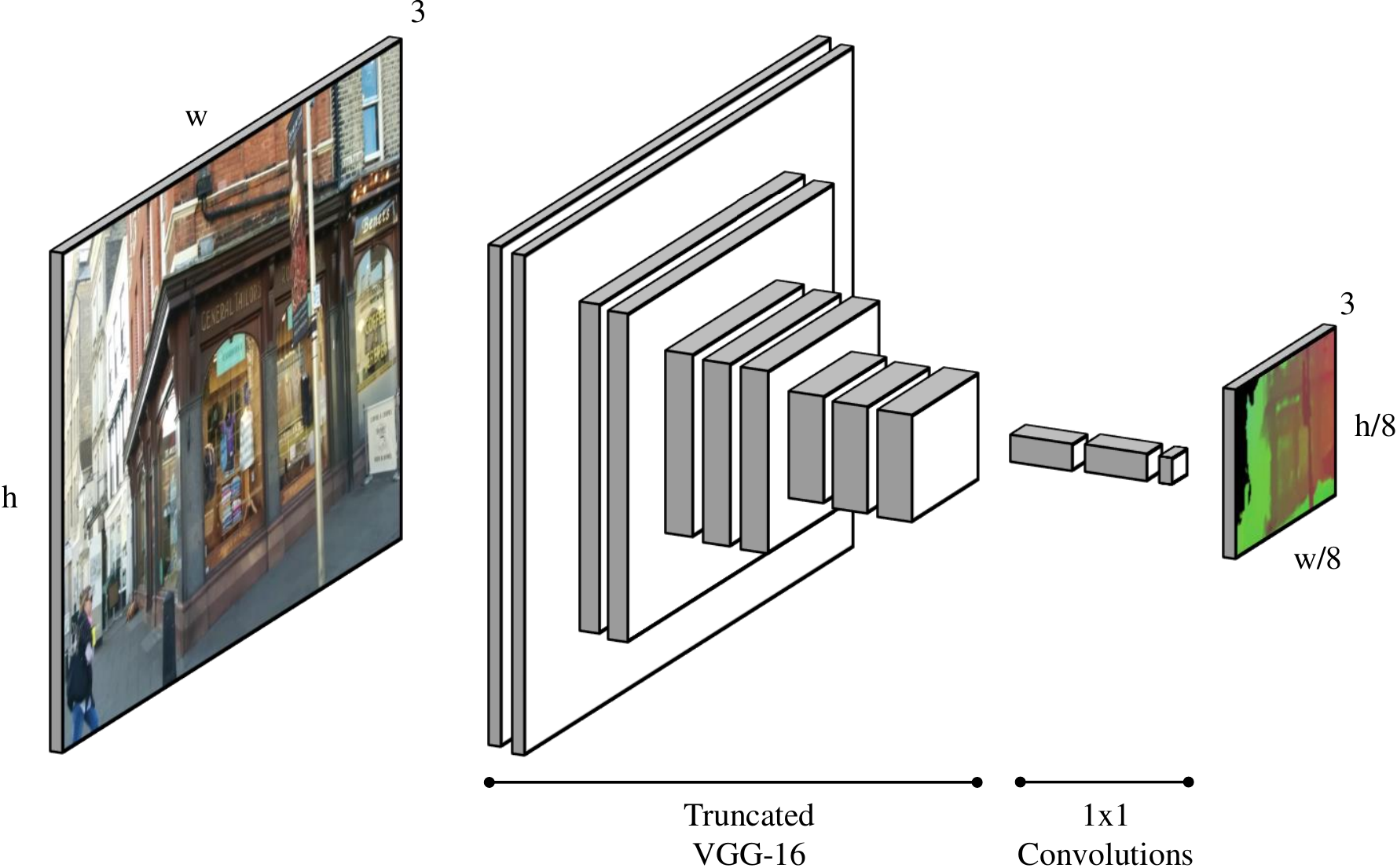}{\textbf{ScoreNet architecture.} We use a truncated VGG-16 feature extractor, followed by several $1 \times 1$ convolutional layers, to regress 3D world space points for a subset of pixels from the original image.}{fig:scorenet}{!t}

\noindent Inspired by Brachmann et al.\ \cite{Brachmann2018CVPR}, we train a ScoreNet with a fully-convolutional, VGG-style \cite{Simonyan2015} architecture to predict correspondences between pixels in the input image and 3D points in world space. Our network takes as input an RGB image of size $w \times h$, and produces as output a $w/8 \times h/8 \times 3$ tensor of 3D world space points corresponding to pixels subsampled densely from the original image on a regular grid with $8$-pixel spacing, i.e.\ pixels $\{(8i,8j) \in [0,w) \times [0,h) : i, j \in \mathbb{N}^+\}$. The architecture (see Figure~\ref{fig:scorenet}) consists of a truncated VGG-16 feature extractor, followed by several $1 \times 1$ convolutional layers to regress a 3D point for each relevant pixel. Each network is trained on the RGB-D training sequence associated with a single scene from one of our datasets (see \S\ref{sec:experiments}). Further details about the architecture and precisely how we train our networks can be found in the supplementary material.

\subsection{Online ScoreNet Prediction Adaptation}
\label{subsec:method-onlineadaptation}

\noindent \textbf{Problem Formulation.} A ScoreNet trained offline on an RGB-D sequence of a scene, as in \S\ref{subsec:method-offlinetraining}, can later be used to relocalise new images in the same scene.
This targets an \emph{offline} formulation of the relocalisation problem, in which both training and testing are performed on the same scene, and there are no constraints on the time available for training. However, this formulation does not take into account the practical requirements on a camera relocaliser for live scenarios such as interactive dense SLAM \cite{Newcombe2011}, in which it is infeasible to spend hours or even days training a relocaliser on the scene of interest; rather, a relocaliser must be trained online as the user moves around the scene, and then be usable immediately when camera tracking fails.

To address such scenarios, we target the alternative \emph{online} formulation of the relocalisation problem proposed by Cavallari et al.\ \cite{Cavallari2017}, in which there are three stages: offline training (`pre-training'), online training and testing. Offline training is performed on sequences of \mbox{RGB-D} frames (with known poses) from one or more scenes, generally other than the target scene. Online training is then performed on a single RGB-D sequence (again with known poses, e.g.\ as produced by a camera tracker) from the target scene. Finally, testing is performed on a single RGB or RGB-D image whose pose is to be determined. (For interactive SLAM, the idea is that a user will move around the scene at online training time, either training a new relocaliser online, or adapting a pre-trained relocaliser online to function in the target scene. If and when camera tracking fails, the trained relocaliser can then be used to recover the camera pose.)

Cavallari et al.\ \cite{Cavallari2017,Cavallari2019} described their online training stage as `adaptation' because they were adapting a pre-trained regression forest to relocalise in the target scene. In particular, they showed that the branching structure of a scene coordinate regression forest can be seen as a scene-independent way of clustering the pixels in an image based on their appearance. Based on this insight, they adapted a pre-trained forest to a new scene by emptying the reservoirs in its leaves and refilling them with points from the new scene at online training time, and then using the forest to look up the reservoirs again to provide correspondences at test time. Inspired by this approach, we show in this paper how to adapt the predictions of a ScoreNet so as to allow these relocalisers too to be deployed in an online context.

\textbf{Reservoir Prediction.} The adaptation scheme described in \cite{Cavallari2017,Cavallari2019} was highly effective, but relied on the fact that their forest does not predict points in any particular scene directly, but instead predicts leaves containing reservoirs of points, which can then be used to generate the needed correspondences. These reservoirs can be refilled with points from the new scene, which is what allowed their method to work, but it is not straightforward to see how it can be transferred to ScoreNets that directly predict individual points in the pre-training scene. To achieve this, we thus propose a new scheme that, rather than clustering pixels into leaves based on routing their associated feature vectors down a regression forest, clusters them into cells in a grid placed over their associated predictions in the pre-training scene (see Figure~\ref{fig:pipeline}). Note that this implicitly clusters pixels in the input image based on their predicted pre-training scene locations, rather than directly based on their appearance.
Intuitively, a ScoreNet, which has been deliberately trained to map similar-looking pixels in an image to similar 3D points in the pre-training scene, can in practice do this for images of any scene, not just the one on which it was trained, and hence pre-training scene location can be used as a reasonable proxy for appearance (see \S\ref{subsec:experiments-correspondencevisualisation} for a discussion).

As mentioned in \S\ref{subsec:method-offlinetraining}, our ScoreNets take an RGB image of size $w \times h$ as input, and produce as output a $w/8 \times h/8 \times 3$ tensor that contains a predicted 3D point (in the scene on which the ScoreNet was trained) for a regularly-spaced subset of pixels in the image. We initially map each of these predicted points, $\mathbf{p} = (p_x, p_y, p_z) \in \mathbb{R}^3$, to a grid cell index as follows. First, we imagine placing a bounded regular cubic grid, with cells of side length $\ell$ and an overall side length of $C\ell$, over the pre-training scene, as shown in Figure~\ref{fig:pipeline}. (The $C$ and $\ell$ values we use can be found in the supplementary material.) Next, for each dimension $k \in \{x,y,z\}$, we compute an index $g(p_k) \in [0 \, .. \, C)$ via
%
\begin{equation}
\textstyle g(p_k) = \mbox{clamp}\left( \left\lfloor \frac{p_k}{\ell} + \frac{C}{2} \right\rceil, \; 0, \; C - 1 \right).
\end{equation}
%
Finally, we combine these three dimension-wise indices into a grid cell index, $G(\mathbf{p})$, via
%
\begin{equation}
\textstyle G(\mathbf{p}) = C^2 g(p_z) + C g(p_y) + g(p_x).
\end{equation}
%
This initial raster-based mapping produces grid cell indices in the range $[0 \, .. \, C^3)$, but in practice, it is undesirable for memory reasons to try to allocate a reservoir for every cell in the grid. Each reservoir may need to store many point clusters, and must be allocated upfront on the GPU with a fixed size. As a result, if every cell in the grid must have a reservoir, then $C$ must be kept small to avoid exceeding the available GPU memory, limiting the size of scene we can handle with our approach.

\stufig{width=\linewidth}{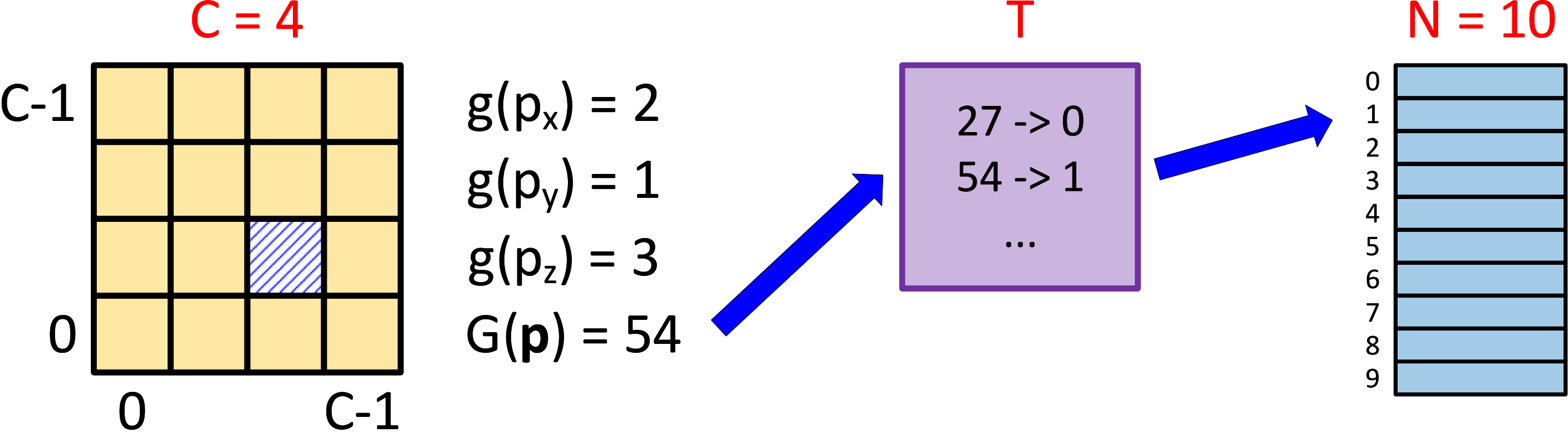}{\textbf{Grid-based reservoir indexing.} Suppose that the 3D point $\mathbf{p}$ that the ScoreNet predicts for a given pixel falls into cell $(2,1,3)$ in a bounded grid placed over the training scene (we show only the $x$ and $y$ dimensions, for simplicity). Then $g(p_x) = 2$, $g(p_y) = 1$ and $g(p_z) = 3$, and we can calculate a grid cell index of $G(\mathbf{p}) = 4^2 \times 3 + 4 \times 1 + 2 = 54$ for $\mathbf{p}$. We use this grid cell index to perform a lookup in a table $T$ that we construct during online training, which stores a mapping from the $C^3$ potential grid cells to a fixed-size buffer of $N \ll C^3$ reservoirs, which we allocate ahead of time. In this case, $T$ maps grid cell $54$ to reservoir $1$, which we then associate with the original pixel.}{fig:reservoirindexing}{!t}

Fortunately, however, there is no need for every grid cell to have a reservoir: as noted by \cite{Niessner2013}, most cells in a scene are empty in practice, and we can exploit this observation to store a sparse set of reservoirs for only those cells that contain predicted points. To achieve this, rather than using the grid cell indices produced as above directly, we instead allocate a fixed-size buffer of $N$ reservoirs upfront, and construct a lookup table $T$ during online training that can be used to map a grid cell index in $[0 \, .. \, C^3)$ to a reservoir index in $[0 \, .. \, N)$: see Figure~\ref{fig:reservoirindexing}. More precisely, we start online training with an empty $T$, and each time we see a grid cell index $G$ for which $T$ has no entry, we add an entry $G \mapsto R$ to $T$ so that $G$ can be remapped to $R \in [0 \, .. \, N)$ in future. We map the first $N$ distinct grid cell indices we see to distinct reservoirs. To handle situations in which the number of grid cells that contain predicted points is greater than $N$, we simply let multiple grid cell indices map to the same reservoir (we randomly pick an existing reservoir to which to map each new grid cell index when no free reservoirs are available). Whilst this can ultimately lead to points with very different appearances being added to the same reservoir, this is not a major problem: as described later in this section, we follow \cite{Cavallari2017} in clustering the points in each reservoir into multiple sets that are disjoint in space, and the only implication of having clusters with different appearances in the same reservoir is that some poor correspondences may be generated. Since the RANSAC-based backend we use \cite{Cavallari2019} is already highly robust to a high proportion of poor correspondences, we would thus expect the practical implications of our reservoir sharing approach to be limited, and indeed our experiment in \S\ref{subsec:experiments-reservoirsharing} shows that this is the case.

\textbf{Reservoir Filling.} The scheme described above allows us to predict reservoir indices in $[0 \, .. \, N)$ for a regularly spaced subset of the pixels in an input image. At online training time, we can use these indices to fill the reservoirs with (world space) points from the target scene. As mentioned above, the online training sequence consists of an ordered set of RGB-D images of the target scene, with their associated $\mathbf{SE}(3)$ poses (which we assume are known, as a result of successfully tracking the camera during online training). For each frame $\mathcal{F}$, we proceed as follows:
\begin{enumerate}[wide]
\item First, we pass the $w \times h$ RGB image through first the ScoreNet and then the grid-based adaptation process just described to produce a reservoir index image of size $w/8 \times h/8$, in which each pixel $(x,y)$ contains the reservoir index to associate with pixel $(8x,8y)$ in the original image.
\item Next, we compute the 3D (world space) point in the target scene corresponding to each pixel $\mathbf{u}$ in the input image for which (i) we have computed a reservoir index and (ii) we have a valid depth value $D(\mathbf{u})$. To do this, we back-project the pixel using the depth to get a point in 3D camera space, and then transform it into world space using the known transformation ${}_\mathcal{W}T_\mathcal{F}$, via
\begin{equation}
\textstyle \mathbf{p} = {}_\mathcal{W}T_\mathcal{F} (D(\mathbf{u})K^{-1}\dot{\mathbf{u}}),
\end{equation}
in which $\dot{\mathbf{u}} = (\mathbf{u}^\top,1)^\top$ is the homogenous form of $\mathbf{u}$, $K$ is the intrinsic calibration matrix for the depth camera, and ${}_\mathcal{W}T_\mathcal{F}$ denotes the transformation from the camera space of frame $\mathcal{F}$ to world space ($\mathcal{W}$). This yields a $w/8 \times h/8 \times 3$ tensor of world space points.
\item Finally, we add each computed world space point to its associated reservoir. We follow \cite{Cavallari2017} in clustering the points we add to each reservoir online using Really Quick Shift (RQS) \cite{Fulkerson2010}, and in maintaining, for each cluster, 3D and colour centroids and a covariance matrix. Since our point clustering is exactly the same as that described in \cite{Cavallari2017}, we refer the reader there for the details of how this works.
\end{enumerate}

\subsection{Camera Pose Estimation}
\label{subsec:method-cameraposeestimation}

\noindent Having filled the reservoirs with clusters of world space points from the target scene at online training time, as per \S\ref{subsec:method-onlineadaptation}, we can then use these clusters at test time to relocalise the camera. To do this, we first pass the $w \times h$ RGB test image through the ScoreNet and the grid-based adaptation process described in \S\ref{subsec:method-onlineadaptation} to produce a reservoir index image of size $w/8 \times h/8$, just as we did during online training. This index image implicitly establishes correspondences between a regularly spaced subset of pixels in the input image and clusters of world space points, which can be used to generate camera pose hypotheses that can be fed to RANSAC. For the actual camera pose estimation, we use the implementation of \cite{Cavallari2019}, which is publicly available in the open-source \emph{SemanticPaint} framework \cite{Golodetz2015SPTR}. Since our contribution in this paper is to the correspondence prediction part of the pipeline, rather than to the camera pose estimation, we summarise how this works only briefly below, and direct the reader to \cite{Cavallari2019} for further details.

\textbf{Hypothesis Generation.} A pose hypothesis $H \in \mathbf{SE}(3)$ maps points in camera space to points in world space. Initially, a large number of pose hypotheses (at most $N_{\max}$) are generated. We follow \cite{Cavallari2019} in generating each pose hypothesis by applying the Kabsch algorithm \cite{Kabsch1976} to $3$ point pairs\footnote{Applying PnP \cite{Hartley2004} to $4$ point pairs instead would make testing RGB-only, but since online training needs depth, Kabsch makes more sense here.} of the form $(\mathbf{x}_i^\mathcal{C},\mathbf{x}_i^\mathcal{W})$, in which $\mathbf{x}_i^\mathcal{C} = D(\mathbf{u}_i)K^{-1}\dot{\mathbf{u}}_i$ is the back-projection of a randomly chosen pixel $\mathbf{u}_i$ in the input image for which a reservoir index has been predicted, and $\mathbf{x}_i^\mathcal{W}$ is a corresponding world space point, randomly sampled from $M(\mathbf{u}_i)$, the modes of the point clusters in the predicted reservoir. We follow \cite{Cavallari2019} in subjecting each generated pose hypothesis to three geometric/colour-based checks, and if one of these checks fails, we try to replace the hypothesis with a new one as described therein.

\textbf{Preemptive RANSAC.} As per \cite{Cavallari2019}, the $\le N_{\max}$ pose hypotheses are first scored and pruned, so that at most $N_{\footnotesize\mbox{cull}}$ hypotheses are retained. The preemptive RANSAC process described in \cite{Cavallari2019} is then used to prune the remaining $\le N_{\footnotesize\mbox{cull}}$ hypotheses down to the best $16$, refining them using Levenberg-Marquardt optimisation \cite{Levenberg1944,Marquardt1963} in the process.

\textbf{Hypothesis Ranking.} Finally \cite{Cavallari2019}, the remaining $16$ hypotheses are refined using ICP \cite{Besl1992} with respect to the 3D scene model, and then scored and ranked by rendering synthetic depth images of the scene model from the ICP-refined poses and comparing these to the live depth image from the camera. The refined pose whose synthetic depth image is most similar to the live depth is then returned as the result.

%% file: text/experiments.tex

\section{Experiments}
\label{sec:experiments}

\noindent In this section, to evaluate our approach, we perform experiments on two well-known relocalisation benchmarks. More experiments can be found in the supplementary material.

\textbf{7-Scenes} \cite{Shotton2013} is a popular RGB-D relocalisation dataset that consists of $7$ different indoor scenes.
Whilst the scenes are relatively small, the captured sequences are in practice quite challenging, exhibiting motion blur, reflective surfaces, and repetitive and/or textureless regions.

\textbf{Cambridge Landmarks} \cite{Kendall2015,Kendall2016,Kendall2017} is an outdoor dataset consisting of $6$ scenes, captured at various locations around Cambridge. It is most commonly used for RGB-only relocalisation, but the coarse 3D SfM models provided for each scene allow it to also be used for RGB-D relocalisation, since it is possible to render depth images of the scene based on these models for each training and testing pose. For our evaluation, we make use of the depth images rendered by Brachmann et al.\ \cite{Brachmann2018CVPR}, to ensure that our results are comparable with those of both their DSAC++ approach and other recent works \cite{Cavallari2019}. As per \cite{Brachmann2018CVPR}, we also compare to a number of other well-known methods that are capable of making use of the 3D model \cite{Brachmann2017CVPR,Kendall2017,Sattler2017}. Note that, in common with other learning-based methods \cite{Brachmann2017CVPR,Brachmann2018CVPR,Brachmann2019}, we ignore the \emph{Street} scene, for which our method too was unable to produce reasonable results (the SfM reconstruction in the dataset appears to be of poor quality for this scene \cite{Brachmann2019}).

\subsection{Relocalisation Performance}
\label{subsec:experiments-performance}

\noindent To evaluate the overall performance of our online relocaliser, and its ability to adapt the predictions of a ScoreNet to a new scene, we pre-trained a ScoreNet for each scene from 7-Scenes \cite{Shotton2013} and Cambridge Landmarks \cite{Kendall2015,Kendall2016,Kendall2017}, and evaluated their performances after grid-based adaptation (see Tables~\ref{tbl:performance-7scenes} and \ref{tbl:performance-cambridge}, and the supplementary material).

Several considerations proved important when training our ScoreNets. 7-Scenes \cite{Shotton2013} unfortunately only contains training and test sequences for each scene (i.e.\ there are no validation sequences), so we used the train/test/validation splits published by \cite{Cavallari2019} when training ScoreNets for these scenes. For Cambridge Landmarks \cite{Kendall2015,Kendall2016,Kendall2017}, the training sequences contain many moving objects (pedestrians, cars, etc.).
For this reason, prior to training, we segmented the training images for each Cambridge Landmarks sequence using an Xception model pre-trained on CityScapes,\footnote{Specifically, the \texttt{xception65\_cityscapes\_trainfine} model from \url{https://github.com/tensorflow/models/blob/master/research/deeplab/g3doc/model_zoo.md}.} and invalidated the depths of all pixels from dynamic classes to avoid them being used during training. We also removed sky and ground pixels, as their depths were unreliable.

Our results on 7-Scenes \cite{Shotton2013} (see Table~\ref{tbl:performance-7scenes}) show that we are able to achieve superior performance to almost all of the methods against which we compared, with the notable exception of the forest-based approach of \cite{Cavallari2019}, which currently outperforms us by around 3\% indoors (although our average median localisation error is the same as that of \cite{Cavallari2019}). However, \cite{Cavallari2019} has the notable downside that whilst it performs well on easier outdoor scenes (e.g.\ see Table~\ref{tbl:performance-cambridge}), it performs extremely poorly on harder scenes such as \emph{Great Court} \cite{Kendall2015,Kendall2016,Kendall2017}.
Indeed, this is still the case even if we explicitly train a forest on \emph{Great Court} itself (see Table~\ref{tbl:performance-cambridge}), and regardless of whether we train a forest with both the RGB and depth features from \cite{Cavallari2019}, or with RGB features alone. By contrast, our approach allows a single ScoreNet trained on \emph{Great Court}, and using only RGB features, to be used to achieve excellent online relocalisation performance not only on all Cambridge Landmarks scenes (see Table~\ref{tbl:performance-cambridge}), but also on all scenes from 7-Scenes (see Table~\ref{tbl:performance-7scenes}).

\newcommand{\best}[1]{\textbf{\textcolor{red}{#1}}}
\newcommand{\secondbest}[1]{\textcolor{blue}{#1}}

\begin{table*}[!t]
\centering
\scriptsize
\begin{tabular}{lcccccccccH}
\toprule
\textsc{Indoor Scenes} & \textbf{Chess}       & \textbf{Fire}        & \textbf{Heads} & \textbf{Office} & \textbf{Pumpkin} & \textbf{Kitchen} & \textbf{Stairs} & \textbf{Average} & \textbf{Avg.\ Med.\ Error} & \textbf{Frame Time (ms)} \\
\midrule
\textbf{RGB-D, online} \\
\midrule
Ours (Office) & 98.95\% & 98.50\% & 99.10\% & 99.78\% & 89.70\% & 84.88\% & 81.60\% & 93.22\% & 0.013m/1.18$^\circ$ \\
Ours (Great Court) & 97.85\% & 97.20\% & 96.80\% & 91.95\% & 84.75\% & 79.86\% & 73.60\% & 88.86\% & 0.019m/1.06$^\circ$ \\
\midrule
Cavallari \emph{et al.} \cite{Cavallari2019} & 99.95\%       & 99.70\%       & 100\%         & 99.48\%              & 90.85\%              & 90.68\%        & 94.20\%       & 96.41\%       & 0.013m/1.17$^\circ$ & 257 \\
\midrule
\textbf{RGB-D, offline} \\
\midrule
Ours (Offline) & 99.80\% & 100\% & N/A & 99.78\% & 90.85\% & 90.36\% & 83.50\% & 94.05\% & 0.015m/1.08$^\circ$ \\
\midrule
Shotton \emph{et al.} \cite{Shotton2013}            & 92.6\% & 82.9\% & 49.4\% & 74.9\% & 73.7\% & 71.8\% & 27.8\% & 67.6\% & -- & -- \\
Guzman-Rivera \emph{et al.} \cite{GuzmanRivera2014} & 96\% & 90\% & 56\% & 92\% & 80\% & 86\% & 55\% & 79.3\% & -- & -- \\
Valentin \emph{et al.} \cite{Valentin2015RF}        & 99.4\% & 94.6\% & 95.9\% & 97.0\% & 85.1\% & 89.3\% & 63.4\% & 89.5\% & -- & -- \\
Brachmann \emph{et al.} \cite{Brachmann2016}        & 99.6\% & 94.0\% & 89.3\% & 93.4\% & 77.6\% & 91.1\% & 71.7\% & 88.1\% & 0.061m/2.7$^\circ$ & -- \\
Meng \emph{et al.} \cite{Meng2018IROS}              & 99.5\% & 97.6\% & 95.5\% & 96.2\% & 81.4\% & 89.3\% & 72.2\% & 90.3\% & 0.017m/0.70$^\circ$ & -- \\
Schmidt \emph{et al.} \cite{Schmidt2017}            & 97.75\% & 96.55\% & 99.8\% & 97.2\% & 81.4\% & 93.4\% & 77.7\% & 92.0\% & -- & -- \\
Brachmann and Rother \cite{Brachmann2018CVPR}       & 97.1\% & 89.6\% & 92.4\% & 86.6\% & 59.0\% & 66.6\% & 29.3\% & 76.1\% & 0.036m/1.1$^\circ$ & -- \\
\midrule
\textbf{RGB-only, offline} \\
\midrule
Brachmann and Rother \cite{Brachmann2018CVPR}       & 93.8\% & 75.6\% & 18.4\% & 75.4\% & 55.9\% & 50.7\% & 2.0\%  & 60.4\% & 0.084m/2.4$^\circ$   & -- \\
Li \emph{et al.} \cite{Li2018ECCV}                 & 96.1\% & 88.6\% & 86.9\% & 80.6\% & 60.3\% & 61.9\% & 11.3\% & 71.8\% & 0.043m/1.3$^\circ$   & -- \\
\bottomrule
\end{tabular}
\vspace{1mm}
\caption{Comparative results on 7-Scenes \cite{Shotton2013} (the \%s are of test frames with $\le 5$cm translation and $\le 5^\circ$ angular errors). We report results after hypothesis ranking (see \S\ref{subsec:method-cameraposeestimation}) for our approach and \cite{Cavallari2019}; see the supplementary material for further results. \emph{Ours (Offline)} denotes a variant in which we pre-trained separate ScoreNets offline for each scene, and then adapted and tested each ScoreNet online on its own scene. The ScoreNets for \emph{Ours (Office)} and \emph{Ours (Great Court)} were pre-trained offline on the bracketed scenes and then adapted and tested online on each different scene.
Note that \emph{Great Court} is a scene from Cambridge Landmarks \cite{Kendall2015,Kendall2016,Kendall2017}, demonstrating the ability of our approach to adapt successfully between datasets.}
\label{tbl:performance-7scenes}
\end{table*}

\newcommand{\bad}{--}

\begin{table*}[!t]
\centering
\scriptsize
\begin{tabular}{llllll}
\toprule
\textsc{Outdoor Scenes} & \textbf{Kings College}       & \textbf{Old Hospital}        & \textbf{Shop Fa\c{c}ade}      & \textbf{St.\ Mary's Church}  & \textbf{Great Court} \\
\midrule
\textbf{Online} \\
\midrule
Ours (Great Court) & 0.011m/0.056$^\circ$ & 0.010m/0.056$^\circ$ & 0.009m/0.040$^\circ$ & 0.011m/0.056$^\circ$ & 0.018m/0.040$^\circ$ \\
                   & \emph{76.97\%} & \emph{66.48\%} & \emph{95.15\%} & \emph{67.17\%} & \emph{77.50\%} \\
\midrule
Cavallari et al.\ \cite{Cavallari2019} \\
-- Office, RGB-D features              & 0.01m/0.06$^\circ$ & 0.01m/0.04$^\circ$ & 0.01m/0.04$^\circ$ & 0.01m/0.06$^\circ$ & $\otimes$ \\
~~~(results from \cite{Cavallari2019}) & \emph{76.97\%} & \emph{82.97\%} & \emph{99.03\%} & \emph{79.62\%} & $\otimes$ \\
-- Great Court, RGB-D features. & \bad{} & \bad{} & 0.009m/0.040$^\circ$ & \bad{} & \bad{} \\
~~~(trained by us)              & \emph{35.86\%} & \emph{41.21\%} & \emph{94.18\%} & \emph{7.55\%} & \emph{0\%} \\
-- Great Court, RGB features & \bad{} & \bad{} & 0.011m/0.056$^\circ$ & \bad{} & \bad{} \\
~~~(trained by us)           & \emph{27.41\%} & \emph{4.40\%} & \emph{74.76\%} & \emph{30.94\%} & \emph{16.58\%} \\
\midrule
\textbf{Offline} \\
\midrule
Ours (Offline) & 0.008m/0.040$^\circ$ & 0.008m/0.040$^\circ$ & 0.009m/0.040$^\circ$ & 0.009m/0.040$^\circ$ & 0.018m/0.040$^\circ$ \\
& \emph{99.71\%} & \emph{100\%} & \emph{100\%} & \emph{99.62\%} & \emph{77.50\%} \\
\midrule
PoseNet (Geom.\ Loss) \cite{Kendall2017}     & 0.99m/1.1$^\circ$ & 2.17m/2.9$^\circ$ & 1.05m/4.0$^\circ$ & 1.49m/3.4$^\circ$  & 7.00m/3.7$^\circ$ \\
Active Search (SIFT) \cite{Sattler2017}      & 0.42m/0.6$^\circ$ & 0.44m/1.0$^\circ$ & 0.12m/0.4$^\circ$ & 0.19m/0.5$^\circ$  & $\otimes$ \\
DSAC (RGB Training) \cite{Brachmann2017CVPR} & *0.30m/0.5$^\circ$ & 0.33m/0.6$^\circ$ & 0.09m/0.4$^\circ$ & *0.55m/1.6$^\circ$ & *2.80m/1.5$^\circ$ \\
DSAC++ \cite{Brachmann2018CVPR}              & 0.18m/0.3$^\circ$  & 0.20m/0.3$^\circ$ & 0.06m/0.3$^\circ$ & 0.13m/0.4$^\circ$  & 0.40m/0.2$^\circ$ \\
\bottomrule
\end{tabular}
\caption{Our average median localisation errors (m/$^\circ$) on Cambridge Landmarks \cite{Kendall2015,Kendall2016,Kendall2017}. Since, like \cite{Cavallari2017,Cavallari2019}, our approach requires depth, we compare to methods that make use of the 3D models provided with the dataset. The \%s (where available) are of test frames with $\le 5$cm translation and $\le 5^\circ$ angular errors.
An $\otimes$ denotes a published failure. A -- is used for the median localisation errors when more than 50\% of the frames failed to relocalise. Numbers marked with a * were the result of end-to-end optimisation that did not converge. For \cite{Cavallari2019}, we report results for forests trained on \emph{Office} and \emph{Great Court}, with both RGB and depth features, and a \emph{Great Court} forest with only RGB features. All forests perform poorly on \emph{Great Court}, with the two forests trained on \emph{Great Court} also performing poorly overall, whereas our online ScoreNet trained on \emph{Great Court} adapts well to all scenes from both this dataset and 7-Scenes \cite{Shotton2013} (see Table~\ref{tbl:performance-7scenes}). See \ref{subsec:experiments-performance} for more discussion.}
\label{tbl:performance-cambridge}
\vspace{-\baselineskip}
\end{table*}

\subsection{Visualising our Approach's Behaviour}
\label{subsec:experiments-correspondencevisualisation}

\begin{stusubfig}{!t}
\begin{subfigure}{.24\linewidth}
\centering
\includegraphics[width=\linewidth]{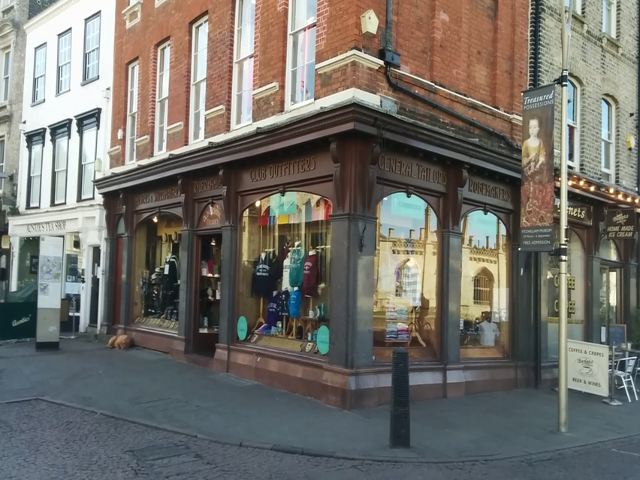}
\end{subfigure}%
\hspace{1mm}%
\begin{subfigure}{.24\linewidth}
\centering
\includegraphics[width=\linewidth]{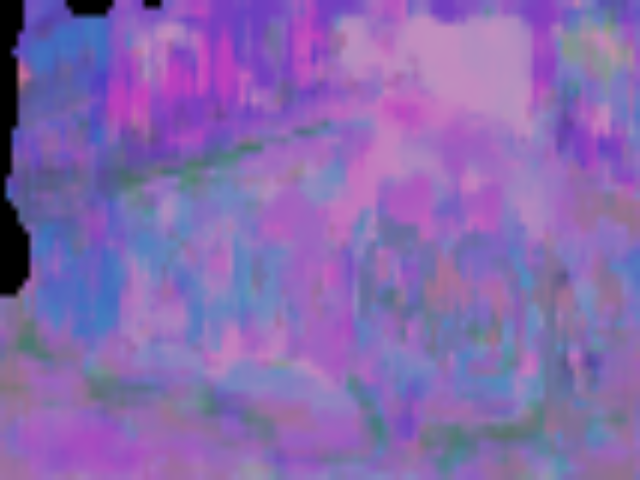}
\end{subfigure}%
\hspace{1mm}%
\begin{subfigure}{.24\linewidth}
\centering
\includegraphics[width=\linewidth]{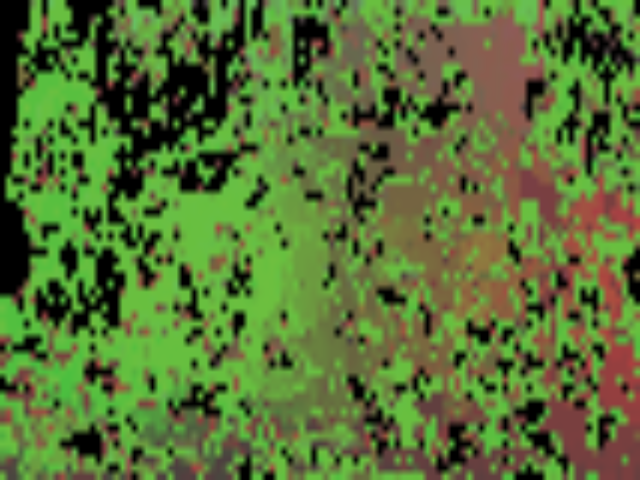}
\end{subfigure}%
\hspace{1mm}%
\begin{subfigure}{.24\linewidth}
\centering
\includegraphics[width=\linewidth]{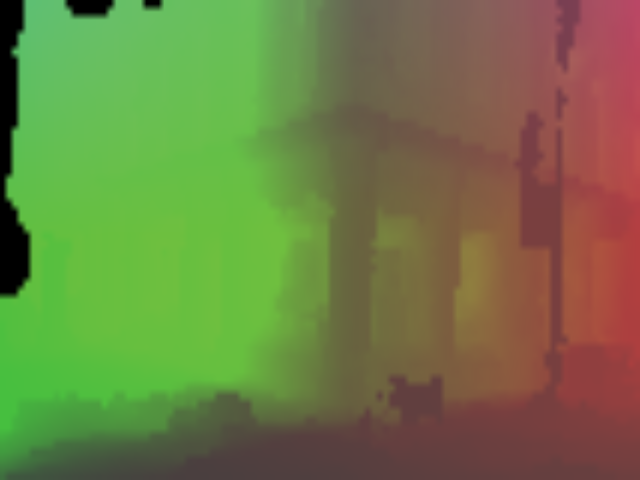}
\end{subfigure}%
\\[0.5mm]
\begin{subfigure}{.24\linewidth}
\centering
\includegraphics[width=\linewidth]{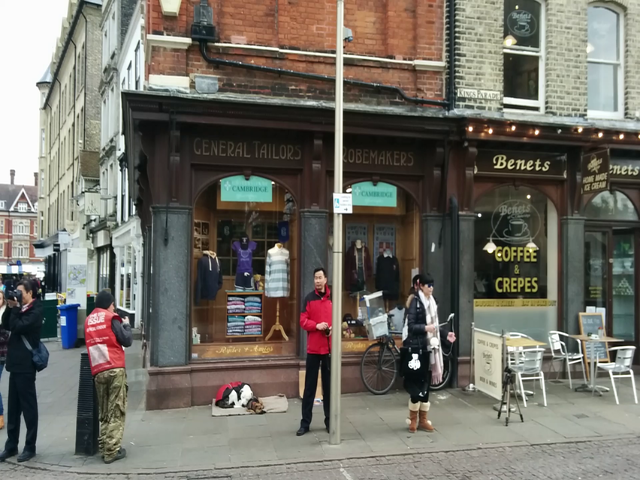}
\end{subfigure}%
\hspace{1mm}%
\begin{subfigure}{.24\linewidth}
\centering
\includegraphics[width=\linewidth]{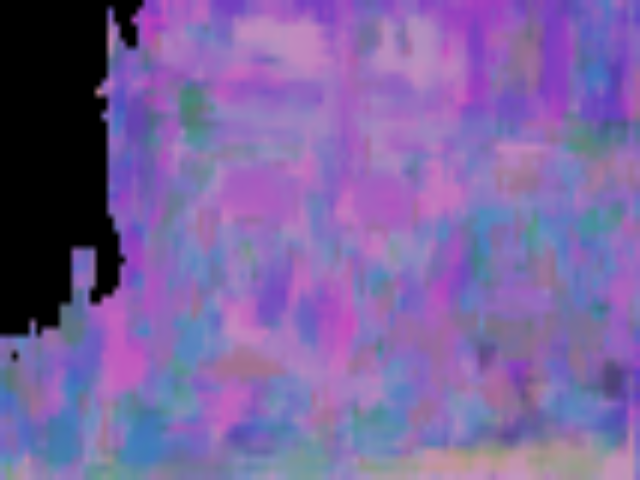}
\end{subfigure}%
\hspace{1mm}%
\begin{subfigure}{.24\linewidth}
\centering
\includegraphics[width=\linewidth]{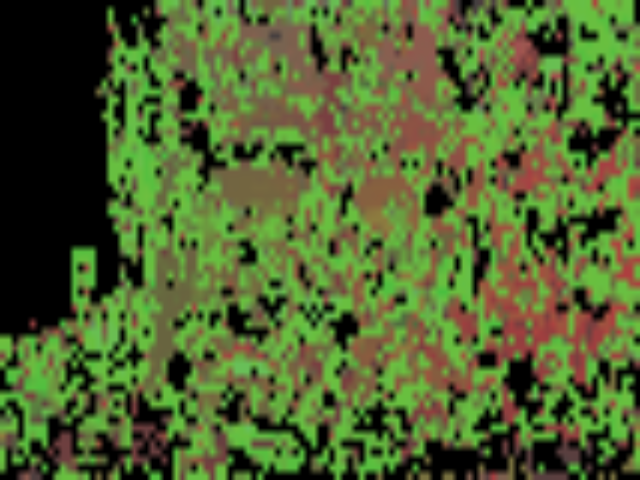}
\end{subfigure}%
\hspace{1mm}%
\begin{subfigure}{.24\linewidth}
\centering
\includegraphics[width=\linewidth]{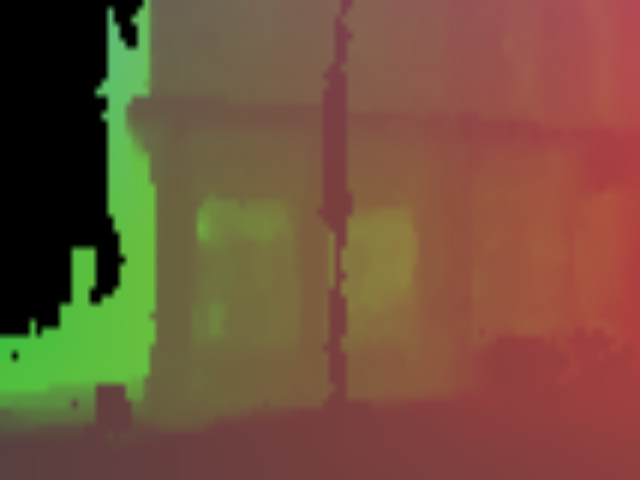}
\end{subfigure}%
\\[0.5mm]
\begin{subfigure}{.24\linewidth}
\centering
\includegraphics[width=\linewidth]{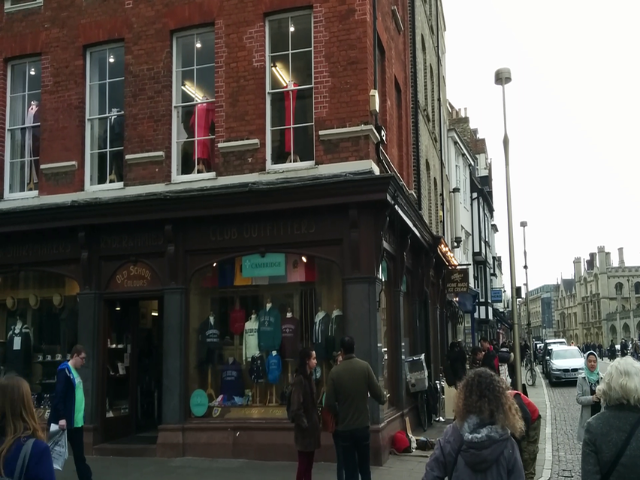}
\end{subfigure}%
\hspace{1mm}%
\begin{subfigure}{.24\linewidth}
\centering
\includegraphics[width=\linewidth]{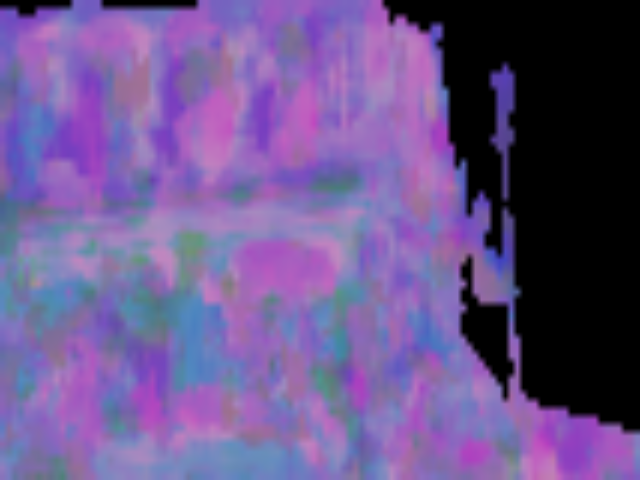}
\end{subfigure}%
\hspace{1mm}%
\begin{subfigure}{.24\linewidth}
\centering
\includegraphics[width=\linewidth]{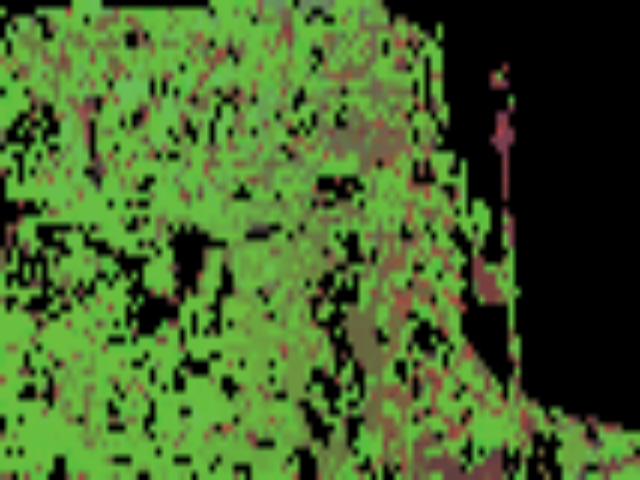}
\end{subfigure}%
\hspace{1mm}%
\begin{subfigure}{.24\linewidth}
\centering
\includegraphics[width=\linewidth]{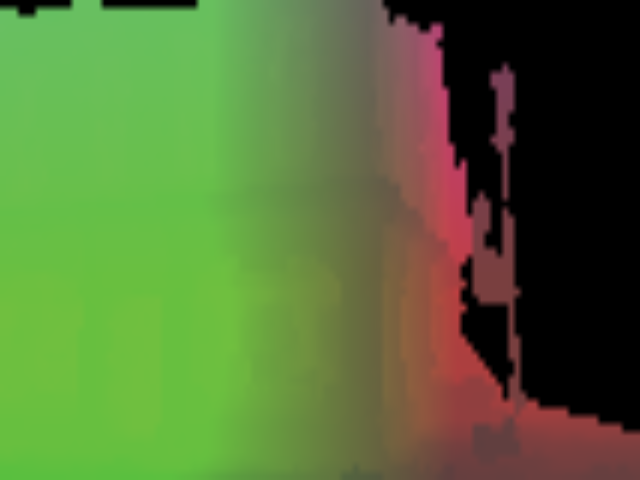}
\end{subfigure}%
\caption{Visualising the raw and adapted points that a ScoreNet trained on \emph{Chess} \cite{Shotton2013} predicts for three images from \emph{Shop Fa\c{c}ade} \cite{Kendall2015,Kendall2016,Kendall2017}, in comparison to the ground truth. Left-to-right: input images, raw predictions (in the \emph{Chess} scene), adapted predictions (in the \emph{Shop Fa\c{c}ade} scene), ground truth points. Points are mapped to scene-specific RGB cubes for visualisation purposes.}
\label{fig:proxy}
\vspace{-\baselineskip}
\end{stusubfig}

\noindent To explain why our approach is able to adapt the predictions of a ScoreNet to enable online relocalisation in a new scene, we visualise the raw and adapted points that a ScoreNet trained on \emph{Chess} \cite{Shotton2013} predicts for three images from \emph{Shop Fa\c{c}ade} \cite{Kendall2015,Kendall2016,Kendall2017}, and compare these to the ground truth (see Figure~\ref{fig:proxy}).
Note how the raw points predicted for similar-looking pixels in the input images are in similar parts of the \emph{Chess} scene (e.g.\ see the green window signs): this is what allows our grid-based adaptation approach to successfully cluster pixels in the input image based on their appearance.
Note also that our approach's performance is also not especially sensitive to the generation of perfect correspondences for every pixel: indeed, as implied by the last two columns of Figure~\ref{fig:proxy}, many predicted correspondences can be incorrect without affecting our ability to relocalise. This is because only $3$ good correspondences are actually needed to successfully estimate the camera pose using the Kabsch algorithm \cite{Kabsch1976}, and so as long as we have predicted enough good correspondences to have a high probability of finding and verifying $3$ good ones during the RANSAC process, our relocaliser is still likely to succeed (see supplementary material). This gives us a significant margin for error when adapting predicted points to a new scene, and makes our approach very robust in practice.

\subsection{Effects of Reservoir Sharing}
\label{subsec:experiments-reservoirsharing}

\noindent To study the impact of reservoir sharing (see \S\ref{subsec:method-onlineadaptation}) on relocalisation performance, we evaluated our relocaliser on 7-Scenes \cite{Shotton2013} for various fixed numbers of reservoirs (i.e.\ values for $N$). The results in Table~\ref{tbl:reservoirsharing} show that our approach is robust to a fairly high level of reservoir sharing: in particular, the performance stays above $90$\% even for values of $N$ as low as $5000$, in a context in which an average of around $29000$ reservoirs are needed if sharing is to be entirely avoided. The performance does eventually decrease for smaller values of $N$, but remains above $80$\% even when only $625$ reservoirs are used. This supports our hypothesis in \S\ref{subsec:method-onlineadaptation} that because the points in each reservoir are clustered into multiple sets that are disjoint in space, and because we use a RANSAC-based backend that is robust even when a high proportion of poor correspondences are generated, our reservoir sharing scheme's overall impact on performance is quite limited in practice for all but extremely low $N$.

\subsection{Timings}
\label{subsec:experiments-timings}

\noindent To better understand the time our approach takes to relocalise a frame, we provide a timing breakdown for our pipeline in Table~\ref{tbl:timings}. Like Cavallari et al.\ \cite{Cavallari2019}, we found that two costly steps were the optimisation of pose hypotheses during RANSAC, and the post-RANSAC ranking of the last $16$ hypotheses. In our case, the initial hypothesis generation is also somewhat costly, since the cost of running a forward pass of the ScoreNet is greater than that of predicting correspondences using a regression forest. Nevertheless, the overall time taken by our relocaliser is only around $292$ms, which is still fast enough to allow our method to be used in live scenarios such as interactive SLAM \cite{Prisacariu2017}.

\begin{table}[!t]
\scriptsize
\centering
\begin{tabular}{cccccccc}
\toprule
N & 625 & 1250 & 2500 & 5000 & 10000 & 20000 & 40000 \\
\% (5cm/5$^\circ$) & 81.21 & 85.31 & 89.07 & 92.02 & 92.95 & 93.34 & 93.22 \\
\bottomrule
\end{tabular}
\caption{The average performance on 7-Scenes \cite{Shotton2013} of our relocaliser trained on \emph{Office} for different numbers of reservoirs $N$ (see \S\ref{subsec:method-onlineadaptation}). On average (over all $7$ scenes), $\approx 29$k reservoirs are needed to avoid reservoir sharing. The performance remains relatively high even when only $5$k reservoirs are used, but eventually decreases for very small $N$.}
\label{tbl:reservoirsharing}
\vspace{-\baselineskip}
\end{table}

\begin{table}[!t]
\scriptsize
\centering
\vspace{2mm}
\begin{tabular}{lc}
\toprule
\textbf{Step} & \textbf{Time (ms)} \\
\midrule
Hypothesis Generation & 72.4 \\
Hypothesis Pruning & 1.2 \\
Inlier Sampling and Energy Computation & 1.4 \\
Optimisation & 72.1 \\
Hypothesis Ranking & 145.2 \\
\midrule
Total & 292.3 \\
\bottomrule
\end{tabular}
\caption{A timing breakdown for the version of our relocaliser trained on \emph{Office} \cite{Shotton2013}. See \S\ref{subsec:experiments-timings} for a discussion.}
\label{tbl:timings}
\vspace{-\baselineskip}
\end{table}

%% file: text/conclusion.tex
\section{Conclusion}

\noindent Visual-only camera relocalisation has received significant attention in recent years because of the key role it plays in a wide variety of computer vision and robotics applications. However, many such applications require a system that can be used online, without expensive prior training on the target scene, for which many state-of-the-art methods, particularly those based on training a network to directly regress the camera pose \cite{Kendall2015}, cannot be used. Of those methods that can be used online, image retrieval methods fail to generalise to poses that are far from the training trajectory, whilst sparse keypoint matching methods tend to struggle in textureless regions, owing to difficulties in detecting suitable keypoints. Scene coordinate regression (SCoRe) methods generalise well to novel poses and can leverage dense correspondences to improve robustness, making them an appealing alternative to such approaches, but hitherto, only the forest-based approach of \cite{Cavallari2017,Cavallari2019} has been able to work online, and that method struggled to generalise to harder outdoor scenes because of its reliance on features that were hand-crafted for indoor use.

In this paper, we have shown how to address this limitation by proposing a way of leveraging the output of a SCoRe network (`ScoreNet') trained on one scene to predict correspondences and relocalise a camera in an entirely different scene. Our approach allows a single ScoreNet, trained on a scene from Cambridge Landmarks \cite{Kendall2015,Kendall2016,Kendall2017} and adapted online, to achieve state-of-the-art performance on all scenes from both an indoor and an outdoor dataset, in under 300ms, without the need for offline training on each individual scene. Notably, unlike the online forest-based approach of \cite{Cavallari2017,Cavallari2019}, which leverages features hand-crafted for indoor use to achieve state-of-the-art results indoors on the 7-Scenes \cite{Shotton2013} and Stanford 4 Scenes \cite{Valentin2016} datasets, but struggles to relocalise well in harder outdoor scenes such as \emph{Great Court}, our method, which uses learnt features, is able to generalise well to such scenes, making it an appealing option for applications that require fast, accurate and online RGB-D camera relocalisation that works equally well in both an indoor and an outdoor context.

%% file: text/supp-arxiv.tex

\section{ScoreNet Architecture}

\lstset{
basicstyle=\scriptsize\ttfamily,
columns=flexible,
breaklines=true
}

\noindent As briefly mentioned in \S\ref{subsec:method-offlinetraining}, the architecture of our ScoreNets consists of a truncated VGG-16~\cite{Simonyan2015} feature extractor, followed by several $1{\times}1$ convolutional layers, to regress a 3D world space point for each relevant pixel (see Figure~\ref{fig:scorenet}).
A more detailed breakdown of the layer structure we use is shown in Figure~\ref{fig:layerstructure}. The feature extraction part of our networks broadly mirrors the structure of a standard VGG-16 feature extractor, but with a couple of modifications to suit our particular context:
\begin{enumerate}
\item Since (i) we want to be able to fill the reservoirs during adaptation at a reasonable rate to ensure that the camera can be relocalised even if camera tracking fails at a relatively early stage, and (ii) we want to predict correspondences for a sizeable subset of the pixels in the input image at test time to give RANSAC more potential correspondences to work with, it is important that the prediction images output by our network are not too small. For this reason, we truncate a conventional VGG-16 feature extractor after three rounds of downsampling, rather than using all of the usual layers.
\item Like \cite{Brachmann2018CVPR}, we use strided convolutions (with a stride of $2$), rather than using max-pooling as in VGG-16.
\end{enumerate}
Our modified feature extractor takes as input an RGB image of size $w \times h$, and produces as output a tensor of size $w/8 \times h/8 \times 512$. This is then fed through a series of $1{\times}1$ convolutional layers, as described in \S\ref{subsec:method-offlinetraining}, to produce a $w/8 \times h/8 \times 3$ tensor of 3D world space points.

To train a ScoreNet, we first initialise the feature extractor with weights learnt by pre-training on ImageNet \cite{Deng2009},\footnote{We adapted the \texttt{torchvision.models.vgg16\_bn} model from TorchVision \cite{TorchVision}, and used the ImageNet-trained weights supplied.} and then train the overall network for the task at hand on the RGB-D training sequence associated with a particular scene in one of our datasets. We use the Adam optimiser~\cite{Kingma2015} with an initial learning rate of $10^{-4}$ (which we reduce by a factor of $10$ whenever the validation loss has not improved over the last $10$ epochs), and batch normalisation~\cite{Ioffe2015}. Training a network (for $160$ epochs) takes from a few hours (for a short training sequence) to a few days (for a longer one).

\begin{figure}[!t]
\begin{lstlisting}
features: Sequential(
(0): Conv2d(3, 64, kernel_size=(3,3), stride=(1,1), padding=(1,1))
(1): BatchNorm2d(64, eps=1e-05, momentum=0.1, affine=True, track_running_stats=True)
(2): ReLU()
(3): Conv2d(64, 64, kernel_size=(3,3), stride=(1,1), padding=(1,1))
(4): BatchNorm2d(64, eps=1e-05, momentum=0.1, affine=True, track_running_stats=True)
(5): ReLU()
(6): Conv2d(64, 128, kernel_size=(3,3), stride=(2,2), padding=(1,1))
(7): BatchNorm2d(128, eps=1e-05, momentum=0.1, affine=True, track_running_stats=True)
(8): ReLU()
(9): Conv2d(128, 128, kernel_size=(3,3), stride=(1,1), padding=(1,1))
(10): BatchNorm2d(128, eps=1e-05, momentum=0.1, affine=True, track_running_stats=True)
(11): ReLU()
(12): Conv2d(128, 256, kernel_size=(3,3), stride=(2,2), padding=(1,1))
(13): BatchNorm2d(256, eps=1e-05, momentum=0.1, affine=True, track_running_stats=True)
(14): ReLU()
(15): Conv2d(256, 256, kernel_size=(3,3), stride=(1,1), padding=(1,1))
(16): BatchNorm2d(256, eps=1e-05, momentum=0.1, affine=True, track_running_stats=True)
(17): ReLU()
(18): Conv2d(256, 256, kernel_size=(3,3), stride=(1,1), padding=(1,1))
(19): BatchNorm2d(256, eps=1e-05, momentum=0.1, affine=True, track_running_stats=True)
(20): ReLU()
(21): Conv2d(256, 512, kernel_size=(3,3), stride=(2,2), padding=(1,1))
(22): BatchNorm2d(512, eps=1e-05, momentum=0.1, affine=True, track_running_stats=True)
(23): ReLU()
(24): Conv2d(512, 512, kernel_size=(3,3), stride=(1,1), padding=(1,1))
(25): BatchNorm2d(512, eps=1e-05, momentum=0.1, affine=True, track_running_stats=True)
(26): ReLU()
(27): Conv2d(512, 512, kernel_size=(3,3), stride=(1,1), padding=(1,1))
(28): BatchNorm2d(512, eps=1e-05, momentum=0.1, affine=True, track_running_stats=True)
(29): ReLU()
)

regression: Sequential(
(0): Conv2d(512, 4096, kernel_size=(1,1), stride=(1,1))
(1): ReLU()
(2): Conv2d(4096, 4096, kernel_size=(1,1), stride=(1,1))
(3): ReLU()
(4): Conv2d(4096, 3, kernel_size=(1,1), stride=(1,1))
)
\end{lstlisting}
\caption{A string representation of the layer structure of our PyTorch-based ScoreNet architecture.}
\label{fig:layerstructure}
\vspace{-\baselineskip}
\end{figure}

Note that it would have been possible to deploy an architecture other than VGG-16 (e.g.\ ResNet-50~\cite{He2016}) for feature extraction purposes. In our case, we chose to use a VGG-based architecture both for simplicity, and for consistency with similar architectures that have been used for this task by other authors \cite{Brachmann2017CVPR,Brachmann2018CVPR}, but there is nothing fundamentally important about this architecture per se: any network that can map pixels that have a similar appearance to similar areas in the pre-training scene should be viable in this context. Indeed, in \S\ref{subsec:trainingefficiency} of this supplementary material, we show that even after a few epochs of training, one of our ScoreNets is already able to predict points in the pre-training scene well enough to allow the camera to be relocalised.

\section{Full Results and Hyperparameter Values}
\label{sec:fullresults}

\begin{table*}[!t]
\centering
\scriptsize
\begin{tabular}{ccc}
\toprule
\textbf{Name} & \textbf{7-Scenes} & \textbf{Cambridge Landmarks} \\
\midrule
clustererSigma & 0.1 & 0.1 \\
clustererTau & 0.05 & 0.4 \\
maxClusterCount & 50 & 50 \\
minClusterSize & 20 & 5 \\
reservoirCapacity & 4096 & 4096 \\
\midrule
maxCandidateGenerationIterations & 6000 & 6000 \\
maxPoseCandidates ($N_{\mbox{max}}$) & 1024 & 2048 \\
maxPoseCandidatesAfterCull ($N_{\mbox{cull}}$) & 64 & 64 \\
maxTranslationErrorForCorrectPose & 0.05 & 0.1 \\
minSquaredDistanceBetweenSampledModes & 0.09 & 0.0225 \\
poseUpdate & True & True \\
ransacInliersPerIteration ($\eta$) & 512 & 512 \\
usePredictionCovarianceForPoseOptimization & True & False \\
\bottomrule
\end{tabular}
\caption{The hyperparameter values we used when testing on 7-Scenes \cite{Shotton2013} and Cambridge Landmarks \cite{Kendall2015,Kendall2016,Kendall2017}.}
\label{tbl:hyperparameters}
\vspace{-\baselineskip}
\end{table*}

\noindent In \S\ref{sec:experiments}, we presented the performance (after ICP and hypothesis ranking) of two variants of our relocaliser: one based on a ScoreNet trained on \emph{Office} from 7-Scenes \cite{Shotton2013}, and another based on a ScoreNet trained on \emph{Great Court} from Cambridge Landmarks \cite{Kendall2015,Kendall2016,Kendall2017}. For space reasons, the results for all other variants were necessarily deferred to this supplementary material, as were the values of the hyperparameters we use in each case. In this section, we present full results (before ICP, after ICP, and after ICP and hypothesis ranking) for variants of our relocaliser trained on every scene from 7-Scenes and Cambridge Landmarks (except \emph{Heads} from 7-Scenes, for which no validation sequence is available). We also provide the raw (i.e.\ \emph{without} grid-based adaptation) results of our ScoreNets on both datasets.

The hyperparameter values we used for each dataset are shown in Table~\ref{tbl:hyperparameters}. Most hyperparameters are the same in both cases, although we use a slightly higher value of \texttt{clustererTau} (the maximum distance there can be between two world space points that are part of the same cluster) and a slightly lower value of \texttt{minClusterSize} (the minimum number of points in a cluster) for the outdoor scenes to reflect the larger scale at which we are operating and the increased difficulty of collecting large enough clusters in that case. We found that these sets of hyperparameters gave good results, although further improvements in performance may be possible via more extensive tuning.

We based the grid cell size ($\ell$) we used for each ScoreNet on the size of the scene on which it was trained. For all the networks trained on sequences from 7-Scenes \cite{Shotton2013}, we used $\ell = 10$cm. For the networks trained on sequences from Cambridge Landmarks \cite{Kendall2015,Kendall2016,Kendall2017}, we used $\ell = 1$m for most scenes, but $\ell = 4$m for \emph{Great Court}, in line with the greater size of that scene ($8000$m$^2$, vs.\ e.g.\ $2000$m$^2$ for \emph{Old Hospital}). In all cases, we set $C$ (the number of grid cells along each side of the grid) so as to ensure that $C\ell$, the side length of the grid, was equal to $1$km. In practice, $C$ can be set to any suitable value that will ensure that the grid covers the pre-training scene.

For the 7-Scenes dataset \cite{Shotton2013}, the raw (i.e.\ \emph{without} grid-based adaptation) results of our 7-Scenes ScoreNets are shown in Table~\ref{tbl:7scenes-raw}, whilst our results with grid-based adaptation enabled are shown in Table~\ref{tbl:7scenes-7scenes-adapted}. Several observations can be made. Firstly, as would be expected given the results reported by \cite{Cavallari2019}, both ICP and hypothesis ranking significantly improve the results in many cases. Secondly, our grid-based adaptation scheme consistently improves relocalisation performance on the scene on which a particular ScoreNet was trained (compare the numbers along the diagonal in Table~\ref{tbl:7scenes-7scenes-adapted} to the raw numbers in Table~\ref{tbl:7scenes-raw}). This is because by predicting a reservoir for each pixel rather than a single point, grid-based adaptation effectively allows us to predict multiple correspondences per pixel (based on the clusters in the reservoirs), which in turn gives RANSAC the opportunity to generate a more diverse range of candidate poses and thereby improve performance. Thirdly, whilst there is some slight variation in performance between ScoreNets trained on different scenes, average performance with adaptation and after ranking for all the networks is state-of-the-art.

For Cambridge Landmarks \cite{Kendall2015,Kendall2016,Kendall2017}, the raw results of our Cambridge ScoreNets are shown in Table~\ref{tbl:cambridge-raw}, whilst our results with grid-based adaptation enabled are shown in Table~\ref{tbl:cambridge-cambridge-adapted}. For outdoor scenes, the 5cm limit on translation error is a much sterner test, which is reflected in the fact that in all cases, the pre-ICP percentages are relatively poor (indeed, most other methods, with the exception of \cite{Cavallari2019}, report only average median localisation errors for this reason). However, as with the 7-Scenes results, it is noticeable that ICP and hypothesis ranking significantly improve performance, leading to state-of-the-art results overall on this dataset. Notably, our grid-based adaptation scheme again improves performance compared to the raw predictions.

Finally, to test how well our grid-based adaptation scheme works between datasets, we evaluated our Cambridge ScoreNets on 7-Scenes (see Table~\ref{tbl:cambridge-7scenes-adapted}), and our 7-Scenes ScoreNets on Cambridge Landmarks (see Table~\ref{tbl:cambridge-7scenes-adapted}). Our Cambridge ScoreNets were able to adapt relatively well to indoor scenes, with each network achieving over $85$\% on average on the 7-Scenes dataset. Conversely, our 7-Scenes ScoreNets also performed well on almost all of the outdoor scenes. They did struggle to adapt well to \emph{Great Court}, which exhibits a much wider range of depth values than the other scenes (see \S\ref{subsec:datasetanalysis}). On the whole, though, these results clearly indicate the potential of our method to adapt well between scenes in different datasets.

\section{Additional Experiments}

\subsection{Generalisation to Novel Poses}

\noindent As noted by \cite{Cavallari2017,Cavallari2019}, an important facet of a relocaliser's performance is its ability to generalise to test images captured from poses that are quite far from the training trajectory.\footnote{Since we are in an online relocalisation context, that means the \emph{online} training trajectory, not the trajectory used during offline pre-training.} To examine how well our relocaliser is able to do this, we follow the methodology originally proposed in \cite{Cavallari2017} of grouping the test images into bins based on the novelty of their ground truth poses with respect to the training trajectory, and evaluating the percentage of frames from each bin that our relocaliser is able to successfully relocalise to within 5cm/5$^\circ$ of the ground truth.

We perform separate evaluations for 7-Scenes \cite{Shotton2013} and Cambridge Landmarks \cite{Kendall2015,Kendall2016,Kendall2017}. For 7-Scenes, we follow \cite{Cavallari2017,Cavallari2019} in using bins specified in terms of a maximum translation and rotation difference with respect to the nearest training pose (a test pose must be within both thresholds to fall into a particular bin), and group all test poses that are more than either 50cm or 50$^\circ$ from the nearest training pose into a single bin.
For Cambridge Landmarks, we found (see \S\ref{subsec:datasetanalysis}) that a wider range of test poses were available: for this reason, we divided the final bin into several new bins that we denote as $< n$ metres, for various $n$. A test pose will fall into such a bin if and only if (i) it could not have fallen into a lower bin, and (ii) its translation distance from the nearest training pose is within the specified threshold.

Our results on 7-Scenes for our relocaliser pre-trained on \emph{Office} are shown in Figure~\ref{fig:generalisation-7scenes}. Although the performance of our relocaliser does gradually decrease as the test pose's distance from the training trajectory increases, it does so quite gracefully, and we remain able to relocalise more than 50\% of frames even at a significant distance ($> 50$cm or $> 50^\circ$) from the training trajectory, indicating that as expected, our approach generalises reasonably well to novel poses.

Similar results on Cambridge Landmarks for our relocaliser pre-trained on \emph{Great Court} can be seen in Figure~\ref{fig:generalisation-cambridge}. As with the indoor scenes, performance gradually decreases as the pose novelty increases, but we remain able to relocalise relatively well even at a distance of several metres from the training trajectory.

\stufig{width=\linewidth}{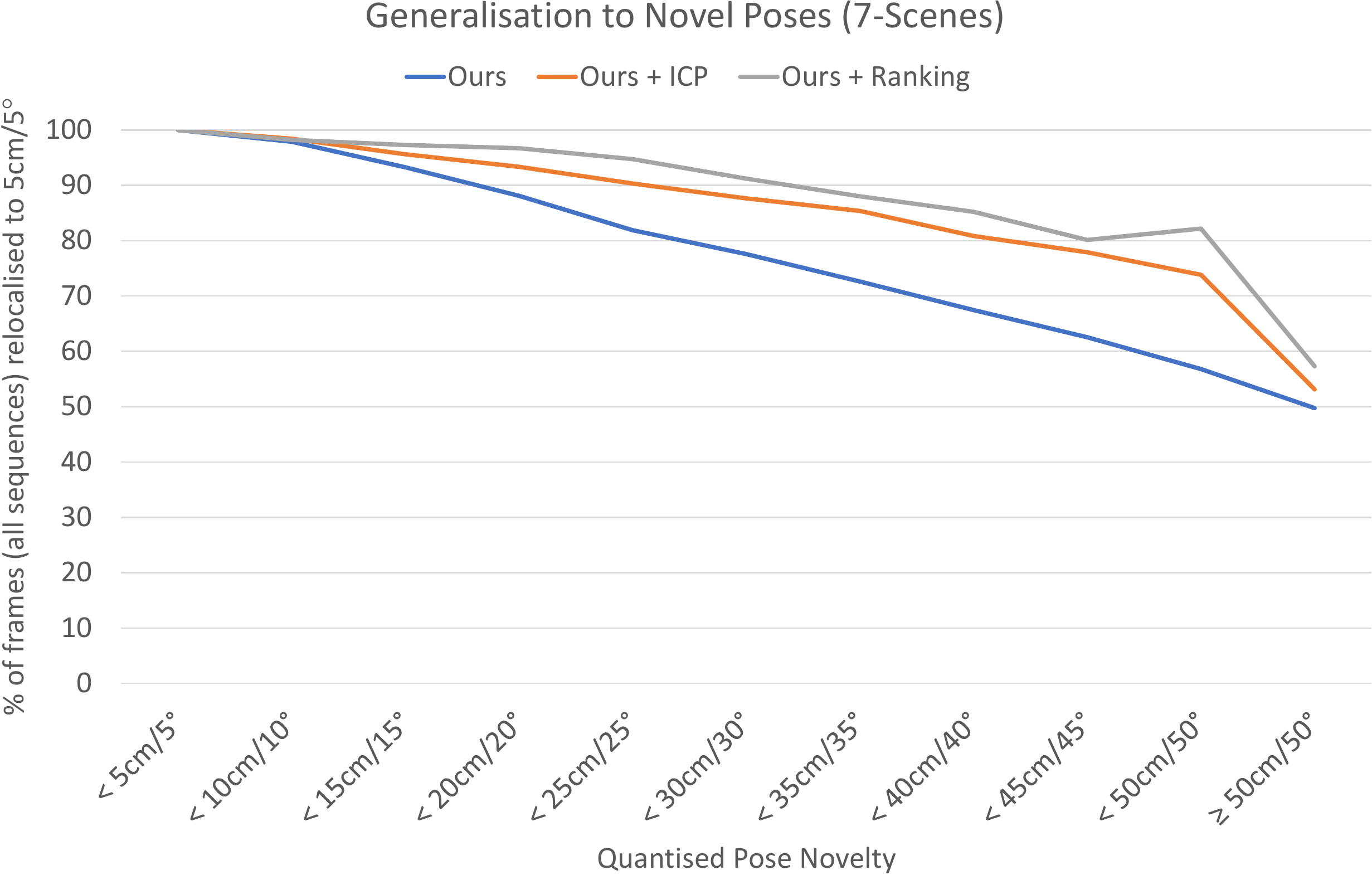}{Evaluating the ability of our relocaliser pre-trained on \emph{Office} to generalise to novel poses (i.e.\ ones far from the online training trajectory) on 7-Scenes \cite{Shotton2013}.}{fig:generalisation-7scenes}{!t}

\stufig{width=\linewidth}{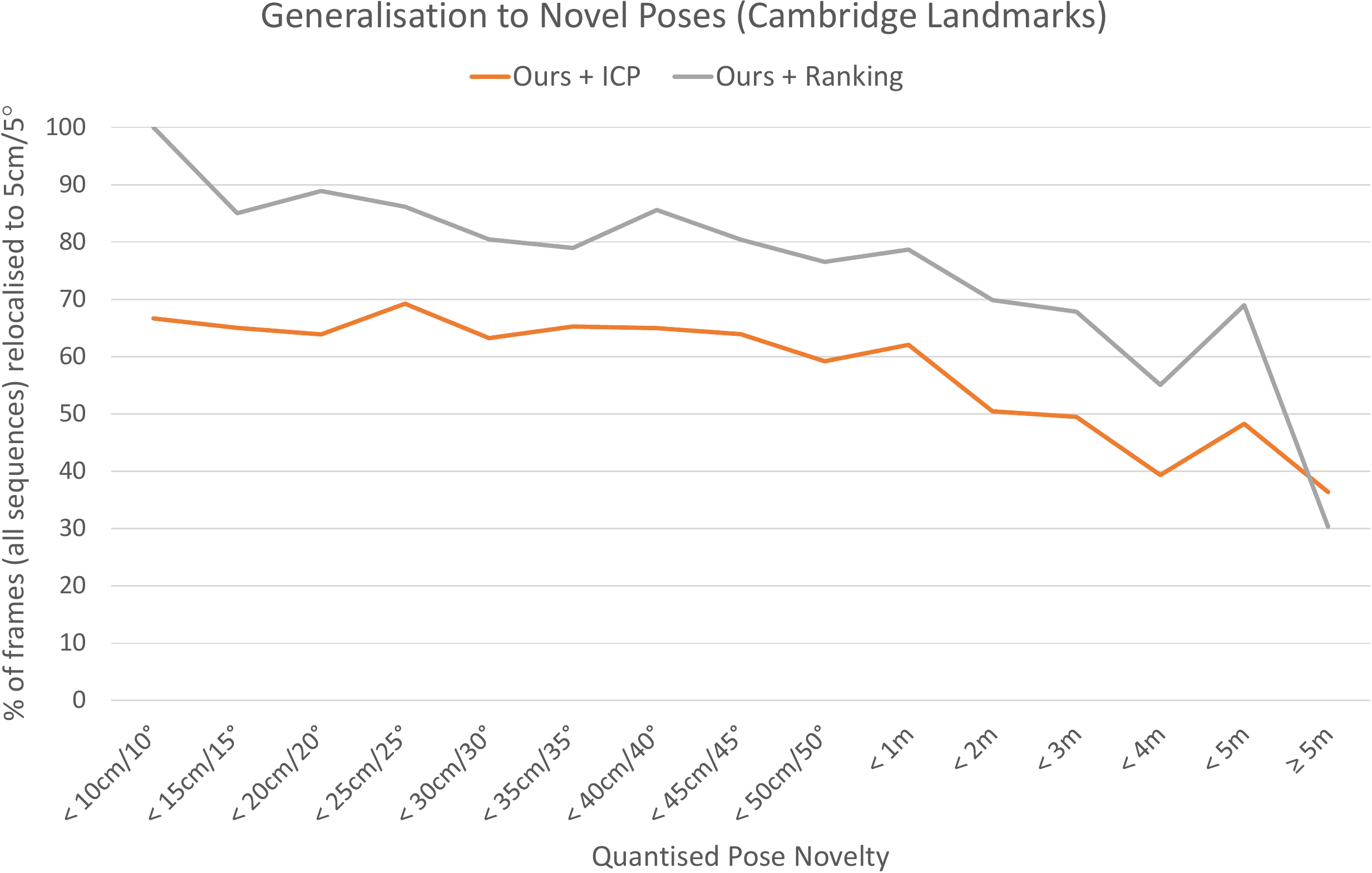}{Evaluating the ability of our relocaliser pre-trained on \emph{Great Court} to generalise to novel poses (i.e.\ ones far from the online training trajectory) on Cambridge Landmarks \cite{Kendall2015,Kendall2016,Kendall2017}. Note that the pre-ICP performance is not shown in this case, since very few images are relocalised to within 5cm/5$^\circ$ before ICP outdoors.}{fig:generalisation-cambridge}{!t}

\subsection{Dataset Analysis}
\label{subsec:datasetanalysis}

\stufig{width=\linewidth}{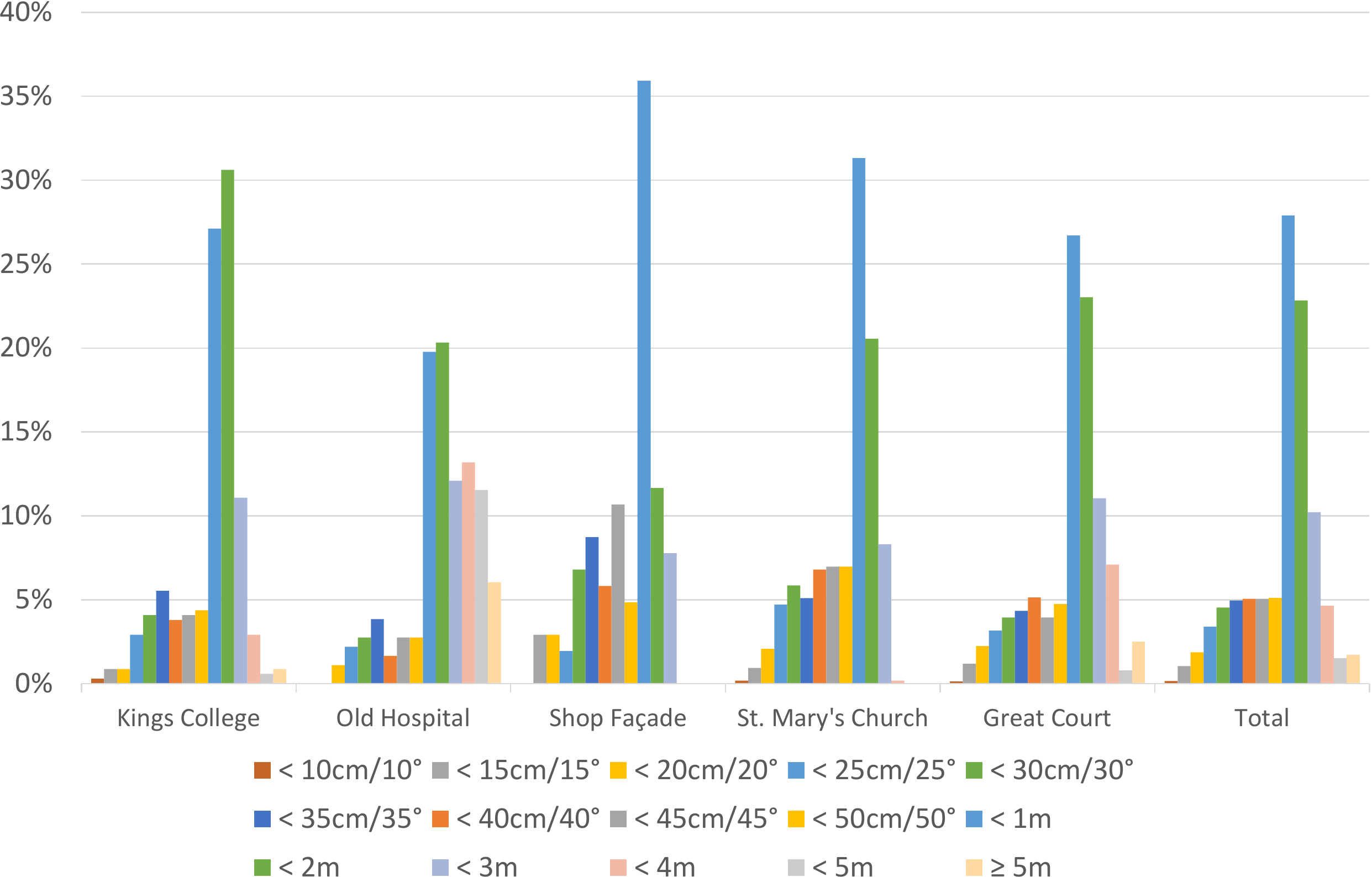}{The proportions of test frames from Cambridge Landmarks \cite{Kendall2015,Kendall2016,Kendall2017} that are within certain distances of the training trajectories. Notably, many of the test poses in this dataset are quite novel, making this in practice a much harder dataset than some of the indoor datasets like 7-Scenes \cite{Shotton2013} and Stanford 4 Scenes \cite{Valentin2016}.}{fig:dataset-cambridge}{!t}

\begin{stusubfig*}{!t}
\begin{subfigure}{.45\linewidth}
\centering
\includegraphics[width=\linewidth]{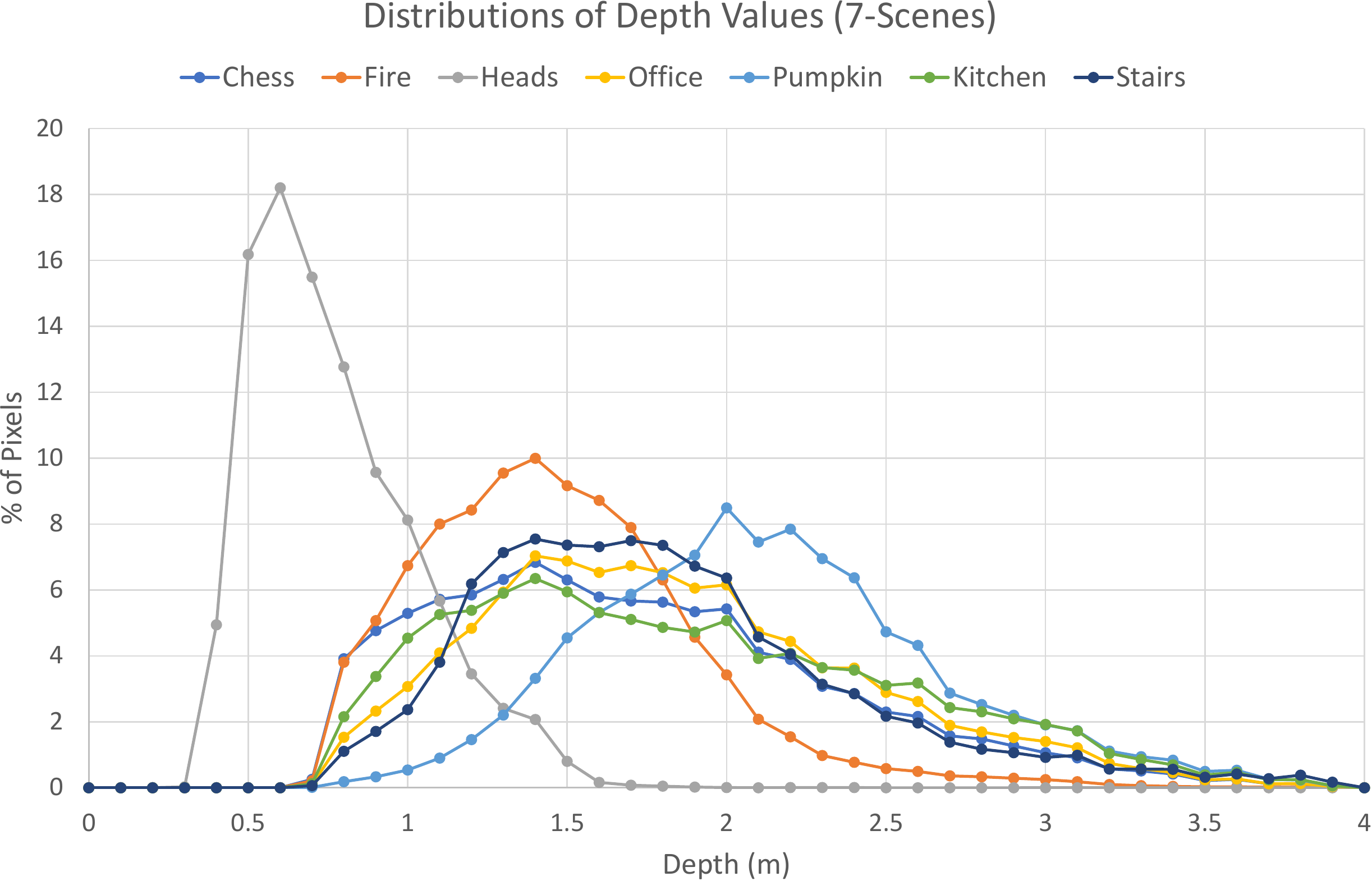}
\end{subfigure}%
\hspace{8mm}%
\begin{subfigure}{.45\linewidth}
\centering
\includegraphics[width=\linewidth]{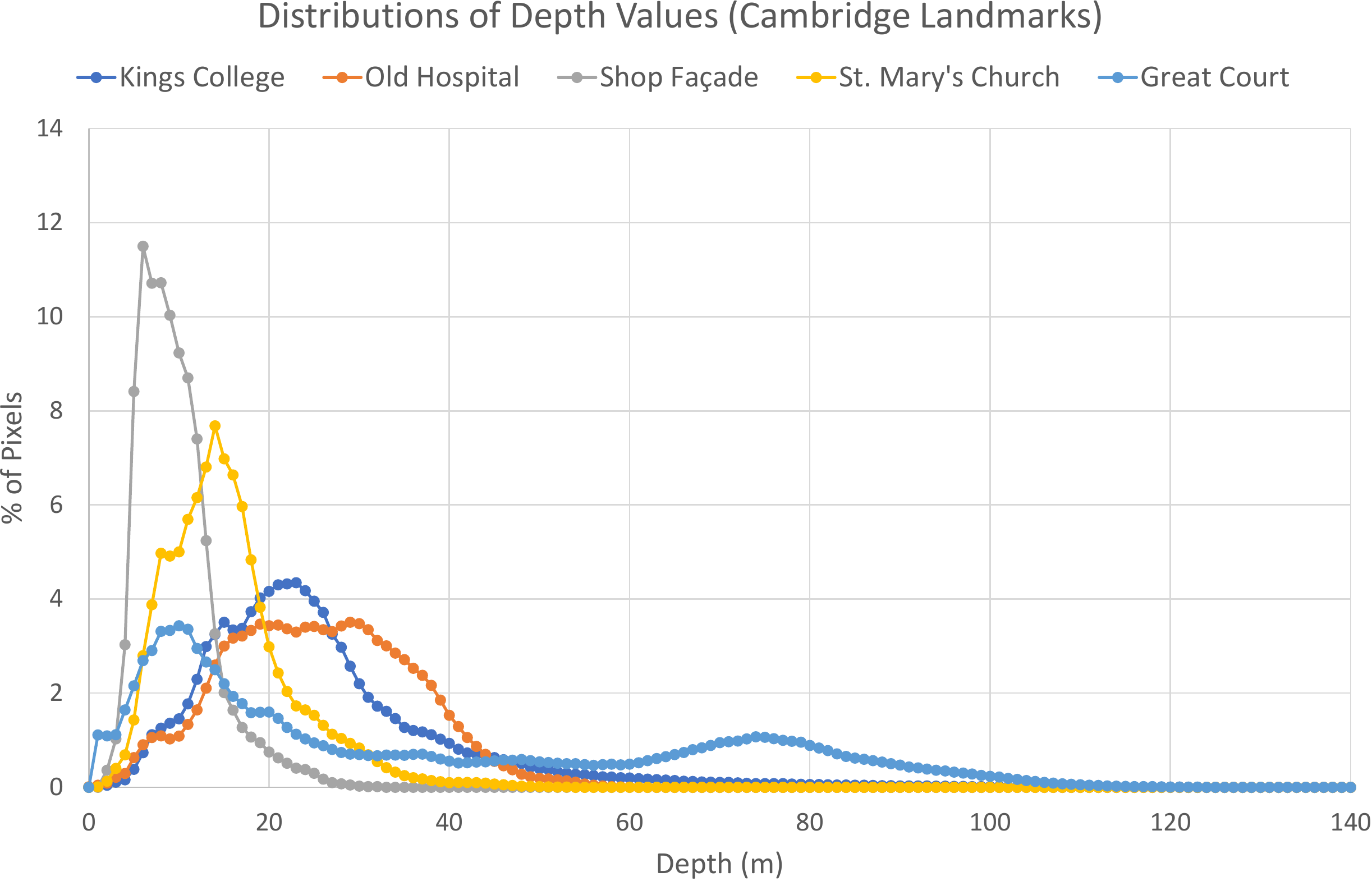}
\end{subfigure}%
\caption{The distributions of depth values in the training sequences of the different scenes in the 7-Scenes \cite{Shotton2013} and Cambridge Landmarks \cite{Kendall2015,Kendall2016,Kendall2017} datasets. Notably, all of the depth values in 7-Scenes are less than 4m, which makes sense given the range of the Kinect sensor with which it was captured, whilst the depth values for the outdoor scenes are much greater.}
\label{fig:depthvalues}
\vspace{-\baselineskip}
\end{stusubfig*}

\noindent In \cite{Cavallari2019}, Cavallari et al.\ provided a detailed analysis of the 7-Scenes \cite{Shotton2013} and Stanford 4 Scenes \cite{Valentin2016} datasets,
to better understand the performance of their relocaliser in each case.
In this section, we provide an analysis of the Cambridge Landmarks \cite{Kendall2015,Kendall2016,Kendall2017} dataset, which we use in this paper to evaluate our relocaliser's outdoor performance.

\textbf{Test Pose Novelty.} The key aspect of the datasets that \cite{Cavallari2019} examined was the percentages of the test poses they contained that were within various distances of the training trajectory. They showed in particular that for Stanford 4 Scenes, the novelty of the test poses is relatively low (most of them are within $30$cm/$30^\circ$ of the training trajectory), whereas for 7-Scenes, far more of the test poses exhibit some novelty (particular those from the \emph{Fire} scene). Nevertheless, the physical constraints of an indoor environment limit how novel the test poses in any such dataset can be in practice (only a small percentage of the test poses in 7-Scenes are more than $50$cm/$50^\circ$ from the training trajectory). For Cambridge Landmarks, we present a similar analysis in Figure~\ref{fig:dataset-cambridge}. Since this is an outdoor dataset, we would naturally expect the range of test poses to be significantly higher than for the indoor datasets, and indeed our results show that this is the case: for every scene (and especially \emph{Old Hospital} and \emph{Great Court}), a large proportion of the test poses are at a significant distance from the training trajectory (more than $50$cm or $50^\circ$), making this dataset quite challenging in practice.

\textbf{Range of Depth Values.} Above and beyond test pose novelty, it is also interesting to examine the range of depth values present in the test frames for the various scenes in the Cambridge Landmarks dataset. Clearly we would expect the depth range to be much larger for outdoor scenes, since on average there are fewer occluders outdoors, and indeed our results in Figure~\ref{fig:depthvalues} show this to be the case. More interestingly, however, it is noticeable that for \emph{Great Court}, the depth range is significantly broader than those for the other scenes. On the one hand, this explains why a relocaliser based on depth-adaptive features that were designed for use indoors with relatively small depth values \cite{Cavallari2017,Cavallari2019} might be expected to perform poorly on such a scene (since large depth values will cause the offsets used to be too small to be usable). On the other hand, it seems plausible that training a ScoreNet on a sequence exhibiting a wider range of depths might yield a network that can predict a more diverse range of points in the pre-training scene, potentially allowing it to generalise better to other scenes at test time (indeed, we observe that our \emph{Great Court} network does in fact have the greatest ability to generalise of all the networks we trained).

\subsection{Correspondence Quality}
\label{subsec:correspondencequality}

\stufig{width=\linewidth}{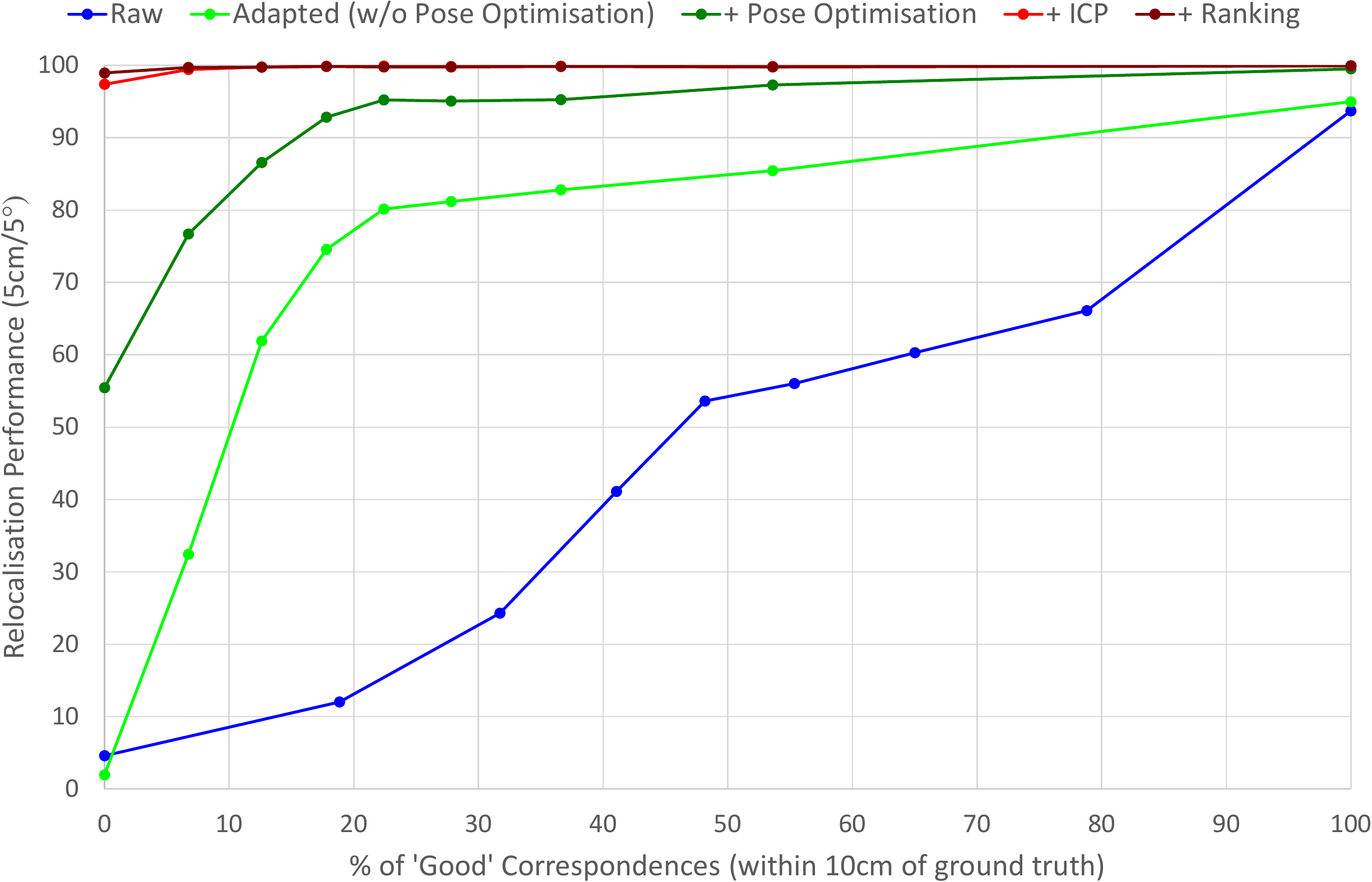}{Evaluating how the performance of our relocaliser trained on \emph{Office} \cite{Shotton2013} and tested on \emph{Office} changes as the percentage of `good' correspondences (those within $10$cm of the ground truth) changes. See \S\ref{subsec:correspondencequality} for a discussion.}{fig:correspondencequality}{!t}

\noindent We observed in \S\ref{subsec:experiments-correspondencevisualisation} that whilst our relocaliser relies on the prediction of correspondences between pixels in the input image and 3D world space points in the target scene to estimate the camera pose, it is not necessary to predict perfect correspondences for every pixel. Indeed, in principle we only need enough good correspondences to allow RANSAC to find and verify $3$ good ones that can be used to estimate the camera pose using the Kabsch algorithm \cite{Kabsch1976}. In this section, we explore the implications of this by evaluating how the relocalisation performance of our \emph{Office} network is affected as we vary the proportion of predicted 3D world space points that lie within $10$cm of the ground truth.

To perform this experiment, we classify, for each frame, all of the predicted 3D world space points\footnote{Note that when using grid-based adaptation, these points will be the modes of clusters in the reservoirs, not points predicted by the ScoreNet.} available as either `good' (within $10$cm of the ground truth) or `poor' (the converse). We then evaluate our relocaliser multiple times on the \emph{Office} test sequence, keeping/discarding different fractions of good/bad correspondences. We further repeat this whole process multiple times for: (i) the raw predictions from the ScoreNet; (ii) the adapted predictions with pose optimisation during RANSAC disabled, no ICP and no hypothesis ranking; (iii) the same as (ii), but with pose optimisation enabled; (iv) the same as (iii), but with ICP enabled; and (v) the same as (iv), but with hypothesis ranking enabled. The reason for exploring the effect that pose optimisation during RANSAC has on the performance is to explain why, with grid-based adaptation enabled, we are able to obtain poses within 5cm/5$^\circ$ of the ground truth for nearly $60$\% of the test frames, even with no `good' correspondences and before ICP or hypothesis ranking. This turns out to be a result of the way in which the third-party camera pose estimation backend we use \cite{Cavallari2019} optimises the pose hypotheses multiple times during the RANSAC process. Without this, our relocaliser performs poorly when there are no `good' correspondences present, as expected.

The results of our experiment are shown in Figure~\ref{fig:correspondencequality}. When grid-based adaptation is disabled, we find that around $50$\% of the correspondences need to be `good' before relocalisation performance starts to become acceptable. By contrast, with grid-based adaptation enabled, a much smaller proportion of `good' correspondences (around $20$\%) are needed. Performing pose optimisation during RANSAC significantly improves the performance when fewer than $20$\% of the correspondences are `good', but does not otherwise change the proportion needed for good performance. Notably, if we then also perform ICP and hypothesis ranking, this further improves the performance so much that excellent relocalisation performance ($> 98$\%) can be achieved even when all of the correspondences are more than $10$cm from the ground truth.

Intuitively, the explanation for this is that the correspondences we use, even though they are individually further than $10$cm from the ground truth, are still good enough that RANSAC with pose optimisation enabled can use them to generate a pose that is within the ICP convergence basin. With hypothesis ranking enabled, there is an even greater chance that one of the $16$ candidate poses generated will be within the ICP convergence basin, allowing relocalisation to succeed. This implies that the link between correspondence quality and relocalisation performance is actually quite weak when the full camera pose estimation backend is used, which in practice means there is a significant margin for error when predicting correspondences. In \S\ref{subsec:trainingefficiency}, we exploit this observation to show that it is possible to significantly reduce the time for which we train our ScoreNets offline without affecting relocalisation performance.

\subsection{Offline Training Efficiency}
\label{subsec:trainingefficiency}

\noindent In \S\ref{subsec:correspondencequality}, we saw that in practice, excellent relocalisation results can be obtained even with a very small proportion of `good' correspondences. Since arguably the only purpose of training a ScoreNet for many epochs is to improve the quality of the correspondences it predicts, and we have just seen that this is not really necessary to achieve excellent relocalisation performance, there are at least two questions that should be asked at this point. Firstly, can even a ScoreNet that has been trained for a much smaller number of epochs than the default (in our case, $160$ epochs) produce correspondences that are good enough for relocalisation to succeed? And secondly, if so, for how many epochs do we really need to train a ScoreNet? To answer these questions, we trained a ScoreNet on the \emph{Great Court} training sequence from Cambridge Landmarks \cite{Kendall2015,Kendall2016,Kendall2017}, and evaluated its post-ranking performance, both with and without grid-based adaptation, on the \emph{Great Court} testing sequence after every $5$ training epochs.

The results of this process are shown in Figure~\ref{fig:training-efficiency}. Notably, they show that it is not in fact necessary to train a ScoreNet for the full $160$ epochs, and that a much smaller number of epochs (around $70$) suffices to achieve maximum relocalisation performance. They also show that with grid-based adaptation enabled, very good results can already be achieved after training for only $20$ epochs, allowing significant time to be saved during the offline training process.

%% file: text/supptables.tex
\providecommand{\best}[1]{\textbf{\textcolor{red}{#1}}}
\providecommand{\secondbest}[1]{\textcolor{blue}{#1}}

\begin{table*}[!p]
\centering
\scriptsize
\begin{tabular}{lcccccc}
\toprule
& \textbf{Chess} & \textbf{Fire} & \textbf{Office} & \textbf{Pumpkin} & \textbf{Kitchen} & \textbf{Stairs} \\
\midrule
Raw & 72.50\% & 41.50\% & 53.38\% & 44.40\% & 39.90\% & 1.20\% \\
& 0.032m/1.495$^\circ$ & 0.061m/2.724$^\circ$ & 0.046m/1.804$^\circ$ & 0.060m/1.865$^\circ$ & 0.068m/2.255$^\circ$ & 0.528m/6.487$^\circ$ \\
\midrule
+ ICP & 98.35\% & 76.65\% & 84.05\% & 74.10\% & 70.90\% & 26.10\% \\
& 0.013m/1.034$^\circ$ & 0.009m/1.043$^\circ$ & 0.010m/1.008$^\circ$ & 0.018m/1.108$^\circ$ & 0.030m/1.480$^\circ$ & 0.552m/1.256$^\circ$ \\
\midrule
+ Ranking & 98.95\% & 88.85\% & 90.80\% & 78.35\% & 79.42\% & 35.40\% \\
& 0.013m/1.032$^\circ$ & 0.008m/1.004$^\circ$ & 0.010m/0.985$^\circ$ & 0.017m/1.096$^\circ$ & 0.027m/1.441$^\circ$ & 0.283m/1.184$^\circ$ \\
\bottomrule
\end{tabular}
\vspace{1mm}
\caption{The raw (i.e.\ \emph{without} grid-based adaptation) results of our ScoreNets on the 7-Scenes dataset \cite{Shotton2013} (the percentages are of test frames with $\le 5$cm translation and $\le 5^\circ$ angular errors, whilst the m/$^\circ$ numbers denote average median localisation errors). A separate ScoreNet was trained on the training sequence for each scene, and tested on the test sequence for that scene. We report three sets of numbers: (i) without ICP or hypothesis ranking, (ii) with ICP but without hypothesis ranking, and (iii) with ICP and hypothesis ranking. As would be expected given the results reported by \cite{Cavallari2019}, both ICP and hypothesis ranking significantly improve the results; however, even after hypothesis ranking, further gains can be achieved using our grid-based adaptation scheme (see Table~\ref{tbl:7scenes-7scenes-adapted}). This is because grid-based adaptation allows us to predict multiple correspondences per pixel (based on the clusters in the reservoirs), which in turn gives RANSAC the opportunity to generate a more diverse range of candidate poses and thereby improve performance.}
\label{tbl:7scenes-raw}
\end{table*}

\begin{table*}[!p]
\centering
\scriptsize
\begin{tabular}{lcccccccc}
\toprule
& \textbf{Chess}       & \textbf{Fire}        & \textbf{Heads} & \textbf{Office} & \textbf{Pumpkin} & \textbf{Kitchen} & \textbf{Stairs} & \textbf{Average} \\
\midrule
\emph{Trained on:} \\ 
\midrule
Chess & 98.65\% & 90.10\% & 93.00\% & 84.50\% & 71.45\% & 84.88\% & 31.50\% & 79.15\% \\
& 0.011m/1.031$^\circ$ & 0.011m/1.049$^\circ$ & 0.010m/1.846$^\circ$ & 0.019m/1.145$^\circ$ & 0.021m/1.161$^\circ$ & 0.022m/1.439$^\circ$ & 0.260m/2.127$^\circ$ & 0.051m/1.400$^\circ$ \\
+ ICP & 99.60\% & 96.90\% & 97.30\% & 94.88\% & 89.00\% & 87.24\% & 49.20\% & 87.73\% \\
& 0.013m/1.033$^\circ$ & 0.007m/0.968$^\circ$ & 0.003m/1.698$^\circ$ & 0.009m/0.957$^\circ$ & 0.016m/1.070$^\circ$ & 0.025m/1.418$^\circ$ & 0.234m/1.099$^\circ$ & 0.044m/1.178$^\circ$ \\
+ Ranking & 99.80\% & 97.60\% & 99.00\% & 97.80\% & 90.15\% & 86.40\% & 78.00\% & 92.68\% \\
& 0.013m/1.031$^\circ$ & 0.008m/0.991$^\circ$ & 0.003m/1.701$^\circ$ & 0.010m/0.960$^\circ$ & 0.016m/1.073$^\circ$ & 0.025m/1.436$^\circ$ & 0.020m/1.056$^\circ$ & 0.014m/1.179$^\circ$ \\
\midrule
Fire & 96.80\% & 89.85\% & 95.70\% & 88.55\% & 73.40\% & 83.68\% & 33.20\% & 80.17\% \\
& 0.010m/1.039$^\circ$ & 0.012m/1.058$^\circ$ & 0.008m/1.823$^\circ$ & 0.017m/1.109$^\circ$ & 0.020m/1.143$^\circ$ & 0.022m/1.444$^\circ$ & 0.260m/1.686$^\circ$ & 0.050m/1.329$^\circ$ \\
+ ICP & 98.55\% & 99.90\% & 98.30\% & 96.60\% & 85.55\% & 87.50\% & 51.10\% & 88.21\% \\
& 0.013m/1.032$^\circ$ & 0.007m/0.962$^\circ$ & 0.003m/1.698$^\circ$ & 0.009m/0.950$^\circ$ & 0.016m/1.070$^\circ$ & 0.025m/1.419$^\circ$ & 0.030m/1.064$^\circ$ & 0.015m/1.171$^\circ$ \\
+ Ranking & 98.70\% & 100.00\% & 99.40\% & 98.05\% & 89.35\% & 85.92\% & 82.30\% & 93.39\% \\
& 0.014m/1.048$^\circ$ & 0.008m/0.970$^\circ$ & 0.003m/1.701$^\circ$ & 0.009m/0.955$^\circ$ & 0.017m/1.073$^\circ$ & 0.025m/1.434$^\circ$ & 0.019m/1.051$^\circ$ & 0.013m/1.176$^\circ$ \\
\midrule
Office & 96.65\% & 88.05\% & 90.80\% & 95.17\% & 67.00\% & 80.02\% & 25.90\% & 77.66\% \\
& 0.010m/1.048$^\circ$ & 0.013m/1.061$^\circ$ & 0.012m/1.948$^\circ$ & 0.013m/1.032$^\circ$ & 0.024m/1.194$^\circ$ & 0.024m/1.470$^\circ$ & 0.311m/2.002$^\circ$ & 0.058m/1.393$^\circ$ \\
+ ICP & 98.80\% & 97.85\% & 97.50\% & 99.88\% & 84.60\% & 85.12\% & 39.10\% & 86.12\% \\
& 0.013m/1.036$^\circ$ & 0.007m/0.969$^\circ$ & 0.003m/1.702$^\circ$ & 0.009m/0.941$^\circ$ & 0.017m/1.074$^\circ$ & 0.026m/1.417$^\circ$ & 0.280m/1.164$^\circ$ & 0.051m/1.186$^\circ$ \\
+ Ranking & 98.95\% & 98.50\% & 99.10\% & 99.78\% & 89.70\% & 84.88\% & 81.60\% & 93.22\% \\
& 0.013m/1.036$^\circ$ & 0.008m/0.985$^\circ$ & 0.003m/1.704$^\circ$ & 0.009m/0.943$^\circ$ & 0.016m/1.071$^\circ$ & 0.025m/1.435$^\circ$ & 0.019m/1.059$^\circ$ & 0.013m/1.176$^\circ$ \\
\midrule
Pumpkin & 96.95\% & 90.90\% & 84.80\% & 86.70\% & 72.65\% & 84.92\% & 26.60\% & 77.65\% \\
& 0.011m/1.072$^\circ$& 0.011m/1.058$^\circ$ & 0.012m/1.872$^\circ$ & 0.019m/1.136$^\circ$ & 0.020m/1.147$^\circ$ & 0.021m/1.435$^\circ$ & 0.281m/2.208$^\circ$ & 0.054m/1.418$^\circ$ \\
+ ICP & 98.05\% & 98.25\% & 91.70\% & 97.22\% & 90.30\% & 87.62\% & 44.20\% & 86.76\% \\
& 0.013m/1.035$^\circ$ & 0.007m/0.965$^\circ$ & 0.003m/1.699$^\circ$ & 0.009m/0.950$^\circ$ & 0.016m/1.066$^\circ$ & 0.025m/1.418$^\circ$ & 0.272m/1.149$^\circ$ & 0.049m/1.183$^\circ$ \\
+ Ranking & 98.60\% & 99.45\% & 95.80\% & 97.68\% & 90.85\% & 87.76\% & 70.90\% & 91.58\% \\
& 0.013m/1.038$^\circ$ & 0.008m/0.977$^\circ$ & 0.003m/1.704$^\circ$ & 0.009m/0.956$^\circ$ & 0.016m/1.069$^\circ$ & 0.025m/1.433$^\circ$ & 0.021m/1.068$^\circ$ & 0.014m/1.178$^\circ$ \\
\midrule
Kitchen & 96.05\% & 85.80\% & 89.00\% & 81.10\% & 69.75\% & 87.20\% & 17.80\% & 75.24\% \\
& 0.012m/1.063$^\circ$ & 0.014m/1.067$^\circ$ & 0.011m/1.904$^\circ$ & 0.022m/1.166$^\circ$ & 0.025m/1.177$^\circ$ & 0.023m/1.427$^\circ$ & 0.322m/2.577$^\circ$ & 0.061m/1.483$^\circ$ \\
+ ICP & 98.35\% & 95.75\% & 92.30\% & 92.47\% & 85.85\% & 90.44\% & 34.70\% & 84.27\% \\
& 0.013m/1.034$^\circ$ & 0.007m/0.967$^\circ$ & 0.003m/1.699$^\circ$ & 0.009m/0.968$^\circ$ & 0.017m/1.070$^\circ$ & 0.024m/1.410$^\circ$ & 0.281m/1.188$^\circ$ & 0.051m/1.191$^\circ$ \\
+ Ranking & 98.80\% & 97.75\% & 96.50\% & 95.62\% & 89.40\% & 90.36\% & 77.30\% & 92.25\% \\
& 0.013m/1.037$^\circ$ & 0.008m/0.979$^\circ$ & 0.003m/1.703$^\circ$ & 0.009m/0.961$^\circ$ & 0.016m/1.072$^\circ$ & 0.024m/1.413$^\circ$ & 0.020m/1.058$^\circ$ & 0.013m/1.175$^\circ$ \\
\midrule
Stairs & 96.70\% & 87.20\% & 87.80\% & 86.75\% & 70.20\% & 79.72\% & 46.20\% & 79.22\% \\
& 0.011m/1.059$^\circ$ & 0.013m/1.059$^\circ$ & 0.011m/1.874$^\circ$ & 0.018m/1.116$^\circ$ & 0.022m/1.193$^\circ$ & 0.027m/1.501$^\circ$ & 0.073m/1.304$^\circ$ & 0.025m/1.301$^\circ$ \\
+ ICP & 98.45\% & 96.85\% & 95.80\% & 96.38\% & 81.65\% & 85.74\% & 56.20\% & 87.30\% \\
& 0.013m/1.033$^\circ$ & 0.007m/0.971$^\circ$ & 0.003m/1.701$^\circ$ & 0.009m/0.952$^\circ$ & 0.017m/1.076$^\circ$ & 0.025m/1.420$^\circ$ & 0.023m/1.053$^\circ$ & 0.014m/1.172$^\circ$ \\
+ Ranking & 98.70\% & 97.80\% & 97.80\% & 97.88\% & 86.25\% & 84.86\% & 83.50\% & 92.40\% \\
& 0.013m/1.043$^\circ$ & 0.008m/0.993$^\circ$ & 0.003m/1.704$^\circ$ & 0.010m/0.957$^\circ$ & 0.016m/1.072$^\circ$ & 0.025m/1.438$^\circ$ & 0.018m/1.037$^\circ$ & 0.013m/1.178$^\circ$ \\
\bottomrule
\end{tabular}
\vspace{1mm}
\caption{The results of our ScoreNets trained on one scene from 7-Scenes \cite{Shotton2013} and subsequently tested on all scenes from 7-Scenes via our grid-based adaptation scheme. The percentages are of test frames with $\le 5$cm translation and $\le 5^\circ$ angular errors, whilst the m/$^\circ$ numbers denote average median localisation errors. A separate ScoreNet was trained on the training sequence for each scene, and subsequently tested on the test sequences for all scenes in the dataset. We report three sets of numbers: (i)~without ICP or hypothesis ranking, (ii) with ICP but without hypothesis ranking, and (iii) with ICP and hypothesis ranking. Compare and contrast with Table~\ref{tbl:7scenes-raw}.}
\label{tbl:7scenes-7scenes-adapted}
\end{table*}

\begin{table*}[!p]
\centering
\scriptsize
\begin{tabular}{lccccc}
\toprule
& \textbf{Kings College} & \textbf{Old Hospital} & \textbf{Shop Fa\c{c}ade} & \textbf{St.\ Mary's Church} & \textbf{Great Court} \\
\midrule
Raw & 2.04\% & 1.10\% & 0.00\% & 0.19\% & 0.13\% \\
& 0.308m/0.588$^\circ$ & 0.514m/0.693$^\circ$ & 0.769m/3.265$^\circ$ & 0.504m/1.797$^\circ$ & 0.785m/0.884$^\circ$ \\
\midrule
+ ICP & 95.04\% & 100\% & 89.32\% & 94.15\% & 52.37\% \\
& 0.009m/0.040$^\circ$ & 0.008m/0.000$^\circ$ & 0.008m/0.040$^\circ$ & 0.009m/0.040$^\circ$ & 0.028m/0.079$^\circ$ \\
\midrule
+ Ranking & 99.13\% & 100\% & 96.12\% & 99.62\% & 79.34\% \\
& 0.008m/0.040$^\circ$ & 0.008m/0.040$^\circ$ & 0.008m/0.056$^\circ$ & 0.009m/0.040$^\circ$ & 0.017m/0.040$^\circ$ \\
\bottomrule
\end{tabular}
\vspace{1mm}
\caption{The raw (i.e.\ \emph{without} grid-based adaptation) results of our ScoreNets on the Cambridge Landmarks dataset \cite{Kendall2015,Kendall2016,Kendall2017} (the percentages are of test frames with $\le 5$cm translation and $\le 5^\circ$ angular errors, whilst the m/$^\circ$ numbers denote average median localisation errors). A separate ScoreNet was trained on the training sequence for each scene, and tested on the test sequence for that scene. We report three sets of numbers: (i) without ICP or hypothesis ranking, (ii) with ICP but without hypothesis ranking, and (iii) with ICP and hypothesis ranking.}
\label{tbl:cambridge-raw}
\end{table*}

\begin{table*}[!p]
\centering
\scriptsize
\begin{tabular}{lcccccc}
\toprule
& \textbf{Kings College}       & \textbf{Old Hospital}        & \textbf{Shop Fa\c{c}ade} & \textbf{St.\ Mary's Church} & \textbf{Great Court} & \textbf{Average} \\
\midrule
\emph{Trained on:} \\ 
\midrule
Kings College & 6.41\% & 0.55\% & 2.91\% & 0.75\% & 0.13\% & 2.15\% \\
& 0.137m/0.444$^\circ$ & 0.955m/2.565$^\circ$ & 0.273m/1.721$^\circ$ & 5.345m/17.904$^\circ$ & 33.042m/85.858$^\circ$ & \\
+ ICP & 97.08\% & 80.22\% & 96.12\% & 48.11\% & 25.26\% & 69.36\% \\
& 0.018m/0.069$^\circ$ & 0.010m/0.056$^\circ$ & 0.014m/0.056$^\circ$ & 5.345m/11.121$^\circ$ & 33.042m/85.858$^\circ$ & \\
+ Ranking & 99.71\% & 85.16\% & 100.00\% & 76.98\% & 31.18\% & 78.61\% \\
& 0.008m/0.040$^\circ$ & 0.009m/0.040$^\circ$ & 0.008m/0.040$^\circ$ & 0.011m/0.056$^\circ$ & 31.972m/81.311$^\circ$ & \\
\midrule
Old Hospital & 0.29\% & 5.49\% & 1.94\% & 0.00\% & 0.13\% & 1.57\% \\
& 1.146m/4.279$^\circ$ & 0.169m/0.407$^\circ$ & 0.308m/2.061$^\circ$ & 2.228m/11.877$^\circ$ & 37.432m/113.246$^\circ$ & \\
+ ICP & 65.31\% & 99.45\% & 95.15\% & 53.21\% & 8.42\% & 64.31\% \\
& 0.013m/0.056$^\circ$ & 0.008m/0.040$^\circ$ & 0.008m/0.056$^\circ$ & 0.021m/0.088$^\circ$ & 37.432m/113.246$^\circ$ & \\
+ Ranking & 74.05\% & 100.00\% & 99.03\% & 75.66\% & 13.82\% & 72.51\% \\
& 0.011m/0.056$^\circ$ & 0.008m/0.040$^\circ$ & 0.008m/0.040$^\circ$ & 0.010m/0.056$^\circ$ & 36.278m/108.150$^\circ$ & \\
\midrule
Shop Fa\c{c}ade & 0.00\% & 0.00\% & 6.80\% & 0.00\% & 0.00\% & 1.36\% \\
& 4.557m/11.934$^\circ$ & 1.533m/3.879$^\circ$ & 0.157m/0.989$^\circ$ & 2.107m/13.161$^\circ$ & 35.936m/108.056$^\circ$ & \\
+ ICP & 46.06\% & 62.64\% & 100.00\% & 56.23\% & 12.24\% & 55.43\% \\
& 3.482m/7.200$^\circ$ & 0.010m/0.069$^\circ$ & 0.008m/0.040$^\circ$ & 0.015m/0.079$^\circ$ & 35.936m/108.056$^\circ$ & \\
+ Ranking & 75.22\% & 76.37\% & 100.00\% & 70.94\% & 14.87\% & 67.48\% \\
& 0.011m/0.056$^\circ$ & 0.009m/0.056$^\circ$ & 0.009m/0.040$^\circ$ & 0.011m/0.056$^\circ$ & 33.772m/102.254$^\circ$ & \\
\midrule
St.\ Mary's Church & 0.87\% & 0.55\% & 1.94\% & 5.66\% & 0.00\% & 1.80\% \\
& 0.679m/2.601$^\circ$ & 0.898m/2.527$^\circ$ & 0.242m/1.666$^\circ$ & 0.181m/0.845$^\circ$ & 39.341m/119.539$^\circ$ & \\
+ ICP & 74.05\% & 79.67\% & 97.09\% & 94.91\% & 9.21\% & 70.99\% \\
& 0.012m/0.056$^\circ$ & 0.009m/0.040$^\circ$ & 0.009m/0.040$^\circ$ & 0.009m/0.040$^\circ$ & 39.341m/119.539$^\circ$ & \\
+ Ranking & 81.63\% & 82.97\% & 99.03\% & 99.62\% & 17.37\% & 76.12\% \\
& 0.011m/0.056$^\circ$ & 0.009m/0.040$^\circ$ & 0.008m/0.040$^\circ$ & 0.009m/0.040$^\circ$ & 37.587m/110.541$^\circ$ & \\
\midrule
Great Court & 0.29\% & 0.00\% & 0.97\% & 0.75\% & 0.79\% & 0.56\% \\
& 0.954m/3.390$^\circ$ & 2.225m/5.216$^\circ$ & 0.481m/3.112$^\circ$ & 11.726m/46.959$^\circ$ & 0.710m/1.671$^\circ$ & \\
+ ICP & 72.89\% & 59.34\% & 77.67\% & 39.25\% & 58.95\% & 61.62\% \\
& 0.011m/0.056$^\circ$ & 0.011m/0.056$^\circ$ & 0.009m/0.056$^\circ$ & 11.726m/46.959$^\circ$ & 0.022m/0.069$^\circ$ & \\
+ Ranking & 76.97\% & 66.48\% & 95.15\% & 67.17\% & 77.50\% & 76.65\% \\
& 0.011m/0.056$^\circ$ & 0.010m/0.056$^\circ$ & 0.009m/0.040$^\circ$ & 0.011m/0.056$^\circ$ & 0.018m/0.040$^\circ$ & \\
\bottomrule
\end{tabular}
\vspace{1mm}
\caption{The results of our ScoreNets trained on one scene from Cambridge Landmarks \cite{Kendall2015,Kendall2016,Kendall2017} and subsequently tested on all scenes from Cambridge Landmarks via our grid-based adaptation scheme. The percentages are of test frames with $\le 5$cm translation and $\le 5^\circ$ angular errors, whilst the m/$^\circ$ numbers denote average median localisation errors. A separate ScoreNet was trained on the training sequence for each scene, and subsequently tested on the test sequences for all scenes in the dataset. We report three sets of numbers: (i)~without ICP or hypothesis ranking, (ii) with ICP but without hypothesis ranking, and (iii) with ICP and hypothesis ranking. Compare and contrast with Table~\ref{tbl:cambridge-raw}.}
\label{tbl:cambridge-cambridge-adapted}
\end{table*}

\begin{table*}[!p]
\centering
\scriptsize
\begin{tabular}{lcccccccc}
\toprule
& \textbf{Chess}       & \textbf{Fire}        & \textbf{Heads} & \textbf{Office} & \textbf{Pumpkin} & \textbf{Kitchen} & \textbf{Stairs} & \textbf{Average} \\
\midrule
\emph{Trained on:} \\ 
\midrule
Kings College & 93.70\% & 81.85\% & 86.50\% & 76.17\% & 60.95\% & 69.76\% & 17.00\% & 69.42\% \\
& 0.012m/1.097$^\circ$ & 0.017m/1.134$^\circ$ & 0.011m/1.917$^\circ$ & 0.026m/1.259$^\circ$ & 0.031m/1.326$^\circ$ & 0.033m/1.588$^\circ$ & 0.352m/3.798$^\circ$ & 0.069m/1.731$^\circ$ \\
+ ICP & 97.05\% & 93.15\% & 90.60\% & 88.07\% & 76.55\% & 81.58\% & 43.00\% & 81.43\% \\
& 0.013m/1.037$^\circ$ & 0.007m/0.971$^\circ$ & 0.003m/1.700$^\circ$ & 0.010m/0.986$^\circ$ & 0.017m/1.087$^\circ$ & 0.026m/1.423$^\circ$ & 0.283m/1.261$^\circ$ & 0.051m/1.209$^\circ$ \\
+ Ranking  & 98.10\% & 96.15\% & 93.50\% & 91.55\% & 81.90\% & 81.20\% & 59.10\% & 85.93\% \\
& 0.013m/1.046$^\circ$ & 0.008m/0.988$^\circ$ & 0.003m/1.702$^\circ$ & 0.010m/0.986$^\circ$ & 0.017m/1.084$^\circ$ & 0.026m/1.441$^\circ$ & 0.024m/1.108$^\circ$ & 0.015m/1.194$^\circ$ \\
\midrule
Old Hospital & 94.70\% & 86.70\% & 90.10\% & 77.68\% & 67.15\% & 76.10\% & 26.10\% & 74.08\% \\
& 0.012m/1.092$^\circ$ & 0.013m/1.067$^\circ$ & 0.009m/1.814$^\circ$ & 0.023m/1.218$^\circ$ & 0.024m/1.234$^\circ$ & 0.027m/1.530$^\circ$ & 0.281m/2.513$^\circ$ & 0.056m/1.495$^\circ$ \\
+ ICP & 97.70\% & 94.35\% & 93.10\% & 90.38\% & 77.20\% & 83.72\% & 48.50\% & 83.56\% \\
& 0.013m/1.034$^\circ$ & 0.007m/0.974$^\circ$ & 0.003m/1.700$^\circ$ & 0.009m/0.972$^\circ$ & 0.017m/1.078$^\circ$ & 0.026m/1.422$^\circ$ & 0.263m/1.093$^\circ$ & 0.048m/1.182$^\circ$ \\
+ Ranking & 98.05\% & 96.90\% & 93.60\% & 93.00\% & 82.30\% & 82.84\% & 75.80\% & 88.93\% \\
& 0.013m/1.039$^\circ$ & 0.008m/0.991$^\circ$ & 0.003m/1.703$^\circ$ & 0.010m/0.976$^\circ$ & 0.017m/1.081$^\circ$ & 0.026m/1.439$^\circ$ & 0.020m/1.056$^\circ$ & 0.014m/1.184$^\circ$ \\
\midrule
Shop Fa\c{c}ade & 93.20\% & 86.90\% & 92.50\% & 77.07\% & 67.85\% & 76.48\% & 24.70\% & 74.10\% \\
& 0.011m/1.080$^\circ$ & 0.012m/1.069$^\circ$ & 0.010m/1.845$^\circ$ & 0.022m/1.196$^\circ$ & 0.025m/1.211$^\circ$ & 0.026m/1.513$^\circ$ & 0.289m/2.405$^\circ$ & 0.056m/1.474$^\circ$ \\
+ ICP & 97.50\% & 96.25\% & 98.40\% & 91.47\% & 81.10\% & 84.64\% & 44.50\% & 84.84\% \\
& 0.013m/1.036$^\circ$ & 0.007m/0.970$^\circ$ & 0.003m/1.700$^\circ$ & 0.009m/0.970$^\circ$ & 0.017m/1.073$^\circ$ & 0.026m/1.420$^\circ$ & 0.272m/1.109$^\circ$ & 0.050m/1.182$^\circ$ \\
+ Ranking & 98.20\% & 98.20\% & 99.30\% & 95.15\% & 85.10\% & 83.48\% & 73.70\% & 90.45\% \\
& 0.014m/1.052$^\circ$ & 0.008m/0.999$^\circ$ & 0.003m/1.705$^\circ$ & 0.010m/0.968$^\circ$ & 0.017m/1.077$^\circ$ & 0.026m/1.443$^\circ$ & 0.021m/1.067$^\circ$ & 0.014m/1.187$^\circ$ \\
\midrule
St.\ Mary's Church & 94.00\% & 76.70\% & 82.10\% & 78.15\% & 55.90\% & 70.60\% & 9.30\% & 66.68\% \\
& 0.013m/1.104$^\circ$ & 0.021m/1.143$^\circ$ & 0.013m/2.024$^\circ$ & 0.024m/1.232$^\circ$ & 0.041m/1.429$^\circ$ & 0.031m/1.563$^\circ$ & 0.575m/5.759$^\circ$ & 0.103m/2.036$^\circ$ \\
+ ICP & 97.35\% & 93.15\% & 87.90\% & 92.43\% & 71.85\% & 80.38\% & 24.20\% & 78.18\% \\
& 0.013m/1.036$^\circ$ & 0.007m/0.975$^\circ$ & 0.003m/1.705$^\circ$ & 0.009m/0.963$^\circ$ & 0.018m/1.105$^\circ$ & 0.027m/1.426$^\circ$ & 0.564m/1.416$^\circ$ & 0.092m/1.232$^\circ$ \\
+ Ranking & 98.30\% & 96.45\% & 93.90\% & 95.65\% & 80.40\% & 81.78\% & 50.20\% & 85.24\% \\
& 0.013m/1.039$^\circ$ & 0.008m/0.993$^\circ$ & 0.003m/1.702$^\circ$ & 0.010m/0.960$^\circ$ & 0.017m/1.085$^\circ$ & 0.026m/1.438$^\circ$ & 0.037m/1.140$^\circ$ & 0.016m/1.194$^\circ$ \\
\midrule
Great Court & 92.50\% & 85.25\% & 88.90\% & 73.70\% & 63.10\% & 66.20\% & 24.40\% & 70.58\% \\
& 0.012m/1.096$^\circ$ & 0.013m/1.082$^\circ$ & 0.011m/1.883$^\circ$ & 0.026m/1.290$^\circ$ & 0.032m/1.318$^\circ$ & 0.035m/1.637$^\circ$ & 0.314m/2.524$^\circ$ & 0.063m/1.547$^\circ$ \\
+ ICP & 96.80\% & 95.10\% & 94.50\% & 87.75\% & 77.25\% & 79.42\% & 41.30\% & 81.73\% \\
& 0.013m/1.038$^\circ$ & 0.007m/0.973$^\circ$ & 0.003m/1.701$^\circ$ & 0.010m/0.984$^\circ$ & 0.017m/1.085$^\circ$ & 0.027m/1.425$^\circ$ & 0.281m/1.185$^\circ$ & 0.051m/1.199$^\circ$ \\
+ Ranking & 97.85\% & 97.20\% & 96.80\% & 91.95\% & 84.75\% & 79.86\% & 73.60\% & 88.86\% \\
& 0.015m/1.060$^\circ$ & 0.008m/0.996$^\circ$ & 0.003m/1.704$^\circ$ & 0.010m/0.985$^\circ$ & 0.018m/1.104$^\circ$ & 0.026m/1.446$^\circ$ & 0.019m/1.056$^\circ$ & 0.014m/1.193$^\circ$ \\
\bottomrule
\end{tabular}
\vspace{1mm}
\caption{The results of our ScoreNets trained on one scene from Cambridge Landmarks \cite{Kendall2015,Kendall2016,Kendall2017} and subsequently tested on all scenes from 7-Scenes \cite{Shotton2013} via our grid-based adaptation scheme. The percentages are of test frames with $\le 5$cm translation and $\le 5^\circ$ angular errors, whilst the m/$^\circ$ numbers denote average median localisation errors. A separate ScoreNet was trained on the training sequence for each scene, and subsequently tested on the test sequences for all scenes in the dataset. We report three sets of numbers: (i)~without ICP or hypothesis ranking, (ii) with ICP but without hypothesis ranking, and (iii) with ICP and hypothesis ranking.}
\label{tbl:cambridge-7scenes-adapted}
\end{table*}

\begin{table*}[!p]
\centering
\scriptsize
\begin{tabular}{lcccccc}
\toprule
& \textbf{Kings College}       & \textbf{Old Hospital}        & \textbf{Shop Fa\c{c}ade} & \textbf{St.\ Mary's Church} & \textbf{Great Court} & \textbf{Average} \\
\midrule
\emph{Trained on:} \\ 
\midrule
Chess & 0.00\% & 0.55\% & 5.83\% & 0.00\% & 0.00\% & 1.28\% \\
& 2.431m/6.882$^\circ$ & 1.463m/3.256$^\circ$ & 0.240m/1.735$^\circ$ & 7.068m/31.042$^\circ$ & 38.155m/113.588$^\circ$ & \\
+ ICP & 51.02\% & 68.13\% & 100.00\% & 43.21\% & 8.68\% & 54.21\% \\
& 0.026m/0.097$^\circ$ & 0.009m/0.056$^\circ$ & 0.008m/0.040$^\circ$ & 7.068m/30.908$^\circ$ & 38.155m/113.588$^\circ$ & \\
+ Ranking & 68.51\% & 77.47\% & 100.00\% & 71.51\% & 13.16\% & 66.13\% \\
& 0.013m/0.069$^\circ$ & 0.009m/0.056$^\circ$ & 0.008m/0.040$^\circ$ & 0.011m/0.056$^\circ$ & 34.747m/105.913$^\circ$ & \\
\midrule
Fire & 0.58\% & 0.00\% & 2.91\% & 0.19\% & 0.00\% & 0.74\% \\
& 1.794m/6.451$^\circ$ & 1.340m/2.977$^\circ$ & 0.282m/1.921$^\circ$ & 2.131m/12.996$^\circ$ & 38.667m/114.043$^\circ$ & \\
+ ICP & 55.69\% & 66.48\% & 98.06\% & 53.21\% & 7.24\% & 56.14\% \\
& 0.017m/0.079$^\circ$ & 0.010m/0.056$^\circ$ & 0.008m/0.040$^\circ$ & 0.019m/0.088$^\circ$ & 38.667m/114.043$^\circ$ & \\
+ Ranking & 77.26\% & 84.62\% & 100.00\% & 73.77\% & 19.61\% & 71.05\% \\
& 0.011m/0.056$^\circ$ & 0.009m/0.040$^\circ$ & 0.008m/0.040$^\circ$ & 0.011m/0.056$^\circ$ & 35.497m/96.146$^\circ$ & \\
\midrule
Office & 0.29\% & 0.55\% & 5.83\% & 0.19\% & 0.00\% & 1.37\% \\
& 5.997m/12.155$^\circ$ & 1.133m/2.906$^\circ$ & 0.222m/1.625$^\circ$ & 1.435m/9.180$^\circ$ & 36.601m/120.281$^\circ$ & \\
+ ICP & 48.98\% & 67.58\% & 94.17\% & 55.47\% & 7.11\% & 54.66\% \\
& 4.215m/11.460$^\circ$ & 0.010m/0.056$^\circ$ & 0.009m/0.040$^\circ$ & 0.016m/0.069$^\circ$ & 36.601m/120.281$^\circ$ & \\
+ Ranking & 65.01\% & 85.71\% & 98.06\% & 78.68\% & 10.79\% & 67.65\% \\
& 0.013m/0.056$^\circ$ & 0.009m/0.040$^\circ$ & 0.009m/0.040$^\circ$ & 0.010m/0.056$^\circ$ & 37.484m/119.759$^\circ$ & \\
\midrule
Pumpkin & 0.87\% & 0.55\% & 2.91\% & 0.38\% & 0.00\% & 0.94\% \\
& 2.490m/6.705$^\circ$ & 1.576m/3.652$^\circ$ & 0.261m/1.977$^\circ$ & 27.319m/99.484$^\circ$ & 42.037m/124.438$^\circ$ & \\
+ ICP & 52.77\% & 65.93\% & 96.12\% & 26.42\% & 3.42\% & 48.93\% \\
& 0.020m/0.079$^\circ$ & 0.010m/0.056$^\circ$ & 0.008m/0.040$^\circ$ & 27.319m/99.484$^\circ$ & 42.037m/124.438$^\circ$ & \\
+ Ranking & 61.22\% & 74.73\% & 98.06\% & 75.85\% & 8.42\% & 63.66\% \\
& 0.013m/0.069$^\circ$ & 0.009m/0.056$^\circ$ & 0.008m/0.040$^\circ$ & 0.010m/0.056$^\circ$ & 41.110m/122.393$^\circ$ & \\
\midrule
Kitchen & 0.00\% & 0.00\% & 2.91\% & 0.38\% & 0.00\% & 0.66\% \\
& 20.956m/40.494$^\circ$ & 1.177m/2.984$^\circ$ & 0.223m/1.589$^\circ$ & 20.186m/79.515$^\circ$ & 41.690m/120.167$^\circ$ & \\
+ ICP & 44.02\% & 64.84\% & 95.15\% & 38.30\% & 7.11\% & 49.88\% \\
& 20.956m/40.494$^\circ$ & 0.010m/0.056$^\circ$ & 0.009m/0.040$^\circ$ & 20.186m/79.515$^\circ$ & 41.690m/120.167$^\circ$ & \\
+ Ranking & 54.81\% & 76.37\% & 99.03\% & 62.83\% & 9.87\% & 60.58\% \\
& 0.018m/0.079$^\circ$ & 0.009m/0.056$^\circ$ & 0.008m/0.040$^\circ$ & 0.013m/0.069$^\circ$ & 40.064m/125.348$^\circ$ & \\
\midrule
Stairs & 0.29\% & 0.55\% & 5.83\% & 0.38\% & 0.00\% & 1.41\% \\
& 7.680m/20.316$^\circ$ & 1.113m/2.779$^\circ$ & 0.250m/2.017$^\circ$ & 1.503m/8.894$^\circ$ & 38.856m/116.153$^\circ$ & \\
+ ICP & 45.48\% & 68.68\% & 89.32\% & 54.34\% & 5.39\% & 52.64\% \\
& 7.019m/20.238$^\circ$ & 0.009m/0.056$^\circ$ & 0.009m/0.056$^\circ$ & 0.016m/0.079$^\circ$ & 38.856m/116.153$^\circ$ & \\
+ Ranking & 67.35\% & 78.02\% & 99.03\% & 70.75\% & 12.50\% & 65.53\% \\
& 0.012m/0.056$^\circ$ & 0.009m/0.040$^\circ$ & 0.009m/0.040$^\circ$ & 0.011m/0.056$^\circ$ & 37.474m/112.747$^\circ$ & \\
\bottomrule
\end{tabular}
\vspace{1mm}
\caption{The results of our ScoreNets trained on one scene from 7-Scenes \cite{Shotton2013} and subsequently tested on all scenes from Cambridge Landmarks \cite{Kendall2015,Kendall2016,Kendall2017} via our grid-based adaptation scheme. The percentages are of test frames with $\le 5$cm translation and $\le 5^\circ$ angular errors, whilst the m/$^\circ$ numbers denote average median localisation errors. A separate ScoreNet was trained on the training sequence for each scene, and subsequently tested on the test sequences for all scenes in the dataset. We report three sets of numbers: (i)~without ICP or hypothesis ranking, (ii) with ICP but without hypothesis ranking, and (iii) with ICP and hypothesis ranking.}
\label{tbl:7scenes-cambridge-adapted}
\end{table*}